\documentclass[runningheads]{llncs}

\usepackage{eccv}

\usepackage{eccvabbrv}

\usepackage{graphicx}
\usepackage{booktabs}
\usepackage{multirow}
\usepackage{wrapfig}
\usepackage[accsupp]{axessibility}  

\usepackage{hyperref}

\usepackage{orcidlink}
\newcommand{\minisection}[1]{\noindent{\textbf{#1}.}}

\begin{document}

\title{Recursive Visual Programming} 

\author{Jiaxin Ge\inst{1} \quad
Sanjay Subramanian\inst{1} \quad
Baifeng Shi\inst{1}\\ \quad Roei Herzig\inst{1} \quad Trevor Darrell\inst{1}}

\authorrunning{Ge et al.}
\institute{UC Berkeley, CA, USA \\
}
\maketitle

\begin{abstract}
Visual Programming (VP) has emerged as a powerful framework for Visual Question Answering (VQA). By generating and executing bespoke code for each question, these methods show advancement in leveraging Large Language Models (LLMs) for complex problem-solving.
Despite their potential, existing VP methods generate all code in a single function, which does not fully utilize LLM's reasoning capacity and the modular adaptability of code. This results in code that is suboptimal in terms of both accuracy and interpretability.
Inspired by human coding practices, we propose {R}ecursive {V}isual {P}rogramming (RVP), which better harnesses the reasoning capacity of LLMs, provides modular code structure between code pieces, and assigns different return types for the sub-problems elegantly. 
RVP approaches VQA tasks with an top-down recursive code generation approach, allowing decomposition of complicated problems into smaller parts. 
We show RVP's efficacy through extensive experiments on benchmarks including VSR, COVR, GQA, and NextQA, underscoring the value of adopting human-like recursive and modular programming techniques for solving VQA tasks. Our code is available at \url{https://github.com/para-lost/RVP}.
\end{abstract}
\section{Introduction}
\label{sec:intro}
\begin{figure}[t!]
    \centering
    \includegraphics[width=12cm]{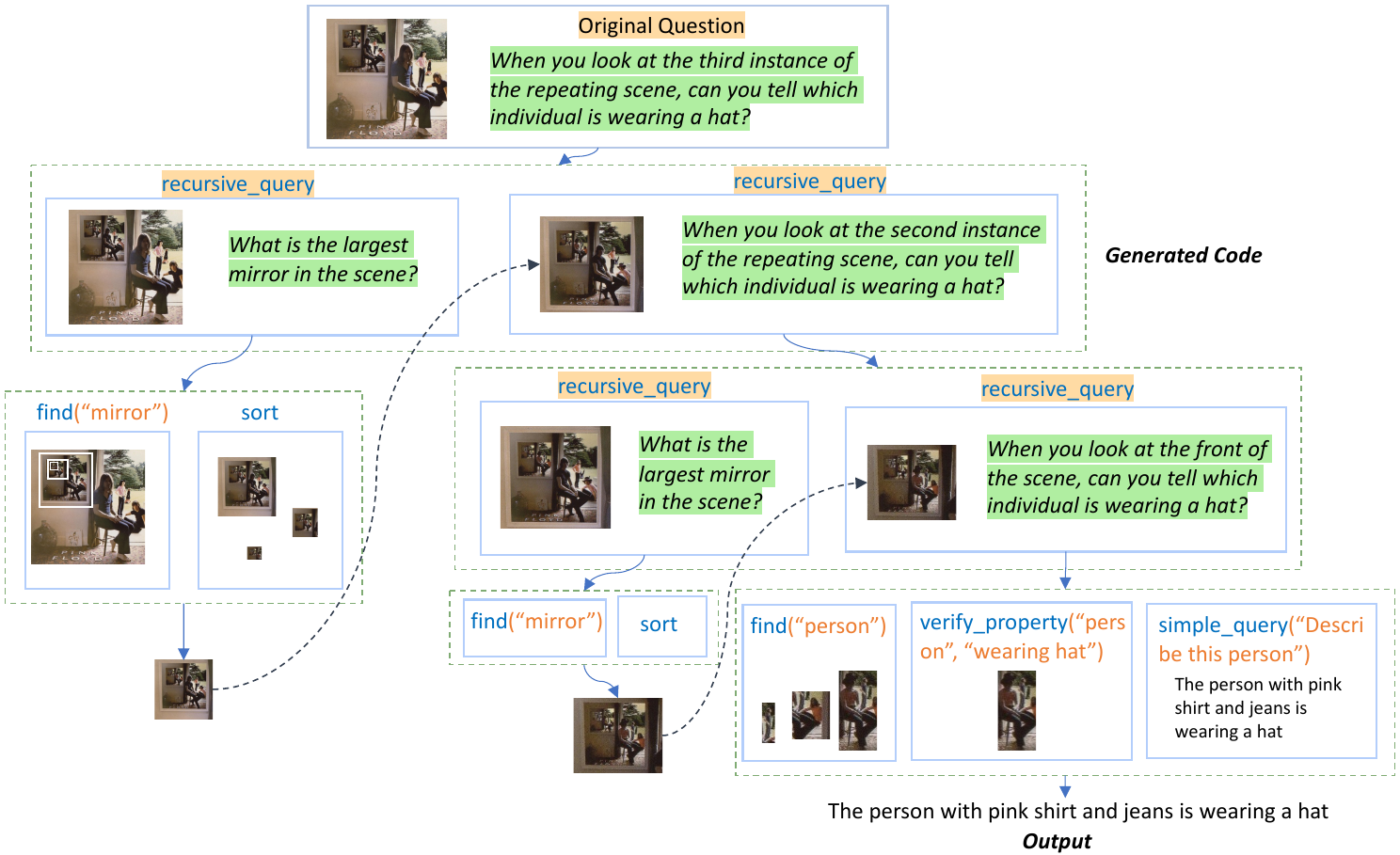}
    \caption{\textbf{A breakdown of Recursive Visual Programming} for a visual question is illustrated with an image where scenes within it contain smaller versions of themselves. To locate a man wearing a hat at the third level of an image, the model first generates code that has two \texttt{recursive\_query} API calls. The initial call identifies the largest mirror (representing the next image level), and the subsequent call seeks ``the man with a hat'' at the second level. The model iteratively generates and executes code through these calls until the final answer is produced without additional recursive calls.}
    \label{fig:model_pipeline_example}
\end{figure}





Visual Question Answering (VQA) lies at the intersection of computer vision and natural language processing, posing the challenge of interpreting visual data to answer questions~\cite{antol2015vqa,Hudson2019GQAAN,Marino2019OKVQAAV}. Following the recent progress on code generation using Large Language Models (LLMs)~\cite{neurips2020gpt3,OpenAI2023GPT4TR,Touvron2023LLaMAOA}, Visual Programming (VP) methods have notably advanced in this area, successfully using LLMs to generate and execute code in few-shot and zero-shot scenarios~\cite{visprog,vipergpt,codevqa}. In these techniques, various vision models are provided to LLM as APIs, and the LLM is used as a planner to utilize these vision models to perform reasoning over images.

Despite these advancements, such planning remains very challenging for Visual Programming. These methods require the model to generate a single code block using the predefined APIs at once. As a result, the model must handle logic, be aware of all details, and utilize the vision APIs appropriately. This has two main drawbacks. First, such approaches do not fully utilize LLM's reasoning capability, which would benefit from breaking down the problem into step-by-step sub-problems, as demonstrated in recent NLP research~\cite{besta2024graph}. Secondly, they fail to utilize the full potential of code because they lack modularity between the code pieces, particularly when the code is one long piece.

We consider how humans tackle these issues. Humans tend to code in a top-down manner: first they break down complex tasks into manageable sub-problems, and then they tackle each through focused code segments. This approach allows the programmer to concentrate on the main logic flow rather than figuring out all the details at once, resulting in a more clear and elegant structure.

\begin{figure}[t!]
    \centering
    \includegraphics[width=12cm]{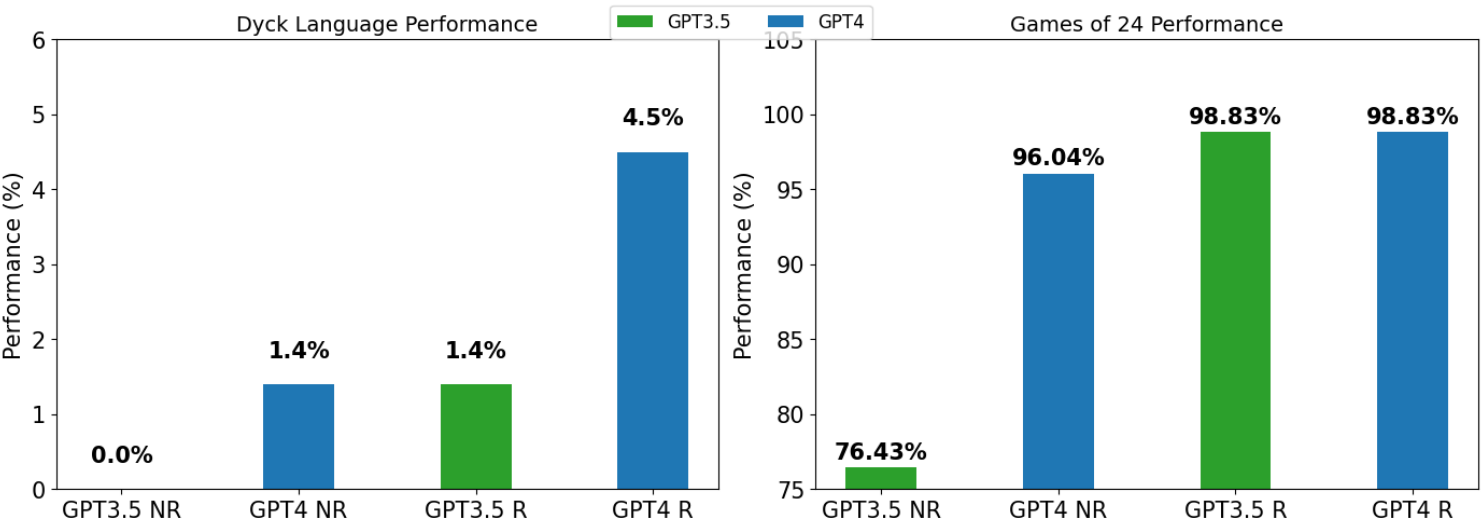}
    \caption{\textbf{Motivating Examples.} Adding the prompt ``solve the problem recursively'' to the input significantly improves the model's performance on traditional coding tasks, Dyck language~\cite{suzgun2022challenging} and Games of 24~\cite{yao2023tree}. ``NR'' refers to non-recursive approach and ``R'' refers to recursive approach. This shows that LLMs can potentially benefit from human's recursive programming approach.}
    \label{fig:motive}
\vspace{-10pt}
\end{figure}

Inspired by this, we hypothesize that incorporating a similar recursive coding strategy would also help solve complex visual question-answering problems. To investigate whether LLMs could indeed benefit from recursive coding, we conduct preliminary experiment of the recursive coding approach on standard symbolic reasoning benchmarks, such as Dyck Language~\cite{suzgun2022challenging} and Games of 24~\cite{yao2023tree}. This is done by simply prompting the LLM to code recursively. We find that utilizing LLMs to code recursively leads to a notable improvement in accuracy, as shown in Figure \ref{fig:motive}. Motivated by this result, we explore in this paper how a recursive coding methodology can be applied to VQA tasks. 

\begin{figure}[t!]
    \centering
    \includegraphics[width=\textwidth]{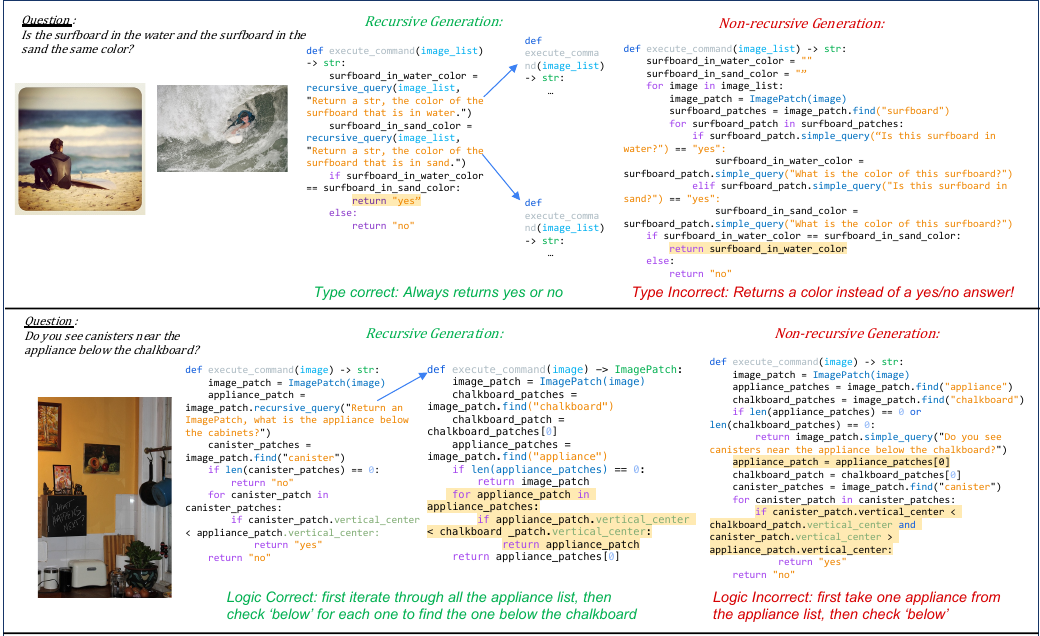}
    \caption{\textbf{Example from COVR (Upper).} Current VP methods fail to provide binary 'yes' or 'no' answers, while Recursive Visual Programming method outputs the correct answer. \textbf{Example from GQA (Middle and Bottom).} Recursive outperforms current non-recursive methods by correctly addressing all details and their associated logic.}
    \label{fig:teaser}
    \vspace{-15pt}
\end{figure}

Specifically, we propose a novel VP method for VQA tasks, named \textbf{R}ecursive \textbf{V}isual \textbf{P}rogramming (\textbf{RVP}). Instead of querying LLM only once to generate a single piece of code, we ask LLM to decompose the question into sub-questions and recursively query itself for each sub-question to generate modular code pieces. For each sub-question, the same process repeats, until reaching an atom level where the current question can be solved by only calling external APIs. See example in Figure \ref{fig:model_pipeline_example}. In contrast to querying LLM only once, we find that using recursive queries to generate code for each sub-question can improve the readability and accuracy of the code, as shown in Figure~\ref{fig:teaser}. Moreover, RVP allows assigning dynamic types to each sub-question, enabling new code pieces to return a variety of types based on the current need. Remarkably, RVP assigns return types unseen in in-context examples, such as \texttt{List[str]}, \texttt{List[ImagePatch]}.

Our contribution can be summarized as follows:
   (i) We introduce a novel recursive approach based on human programming. This approach is a top-down programming method that better harnesses LLM's reasoning capacity and enhances modularity between multiple code pieces. It is easily applicable to any existing VP model for visual question answering task. (ii) We allow dynamic return types for the sub-problems, which enhances flexibility and adaptability. (iii) We demonstrate improved performance on several standard benchmarks, such as GQA, VSR, NextQA and COVR, demonstrating the effectiveness of our approach. We also show better interpretability than non-recursive VP methods.

\section{Related Work}
\textbf{Modular Visual Reasoning.}
There is a long line of work that seeks to combine modularity with the expressivity of deep neural networks. Early work on this topic converts the output of a parser \cite{nmn} to a program or trains a program generator either separately from or jointly with the execution modules \cite{andreas2016,hu2018}. Later work finds that such models are often challenging to train on real-world datasets, particularly when interpretability of individual module outputs is expected \cite{obtaining2020}. Other work improves performance on multi-hop visual reasoning without explicitly defining a set of modules, e.g., by predicting and operating over a scene graph \cite{hudson2019}.\\
\textbf{Visual Programming.}
Recent language models can accurately generate code from text descriptions when prompted with a small number of examples \cite{codex}. This advance has enabled a few-shot modular visual reasoning approach, commonly called visual programming \cite{visprog}, which consists of prompting a language model for a program and executing the program with pre-trained vision models. This paradigm has led to impressive results in visual question answering, video question answering, visual and tabular question answering, referring expression comprehension, and text-based image editing \cite{binder,codevqa,vipergpt,vpgen,chameleon}. Programming with LLMs has also been applied to produce plans and reward functions for robots \cite{codeAsPolicies,languageToRewards,eureka}. Our work builds on these techniques by enabling the LLM to generate code in multiple steps--first planning the overall process and then filling in the details--rather than all at once.\\
\textbf{Language Models for Reasoning Tasks.}
In the text domain, there is a large body of work on using language models for reasoning tasks. In particular, several papers have shown the efficacy of code generation as a scaffold for such tasks \cite{code4struct,madaan2022Language,gao2022Pal,chen2022Program}. Aside from code generation, prior work has also shown that decomposing complex queries into simpler ones improves the performance of language models \cite{leasttomost,selfask}. Our RVP approach applies this insight to improve the code generation component of visual programming.
\begin{figure}[h]
    \centering
    \includegraphics[width=12cm]{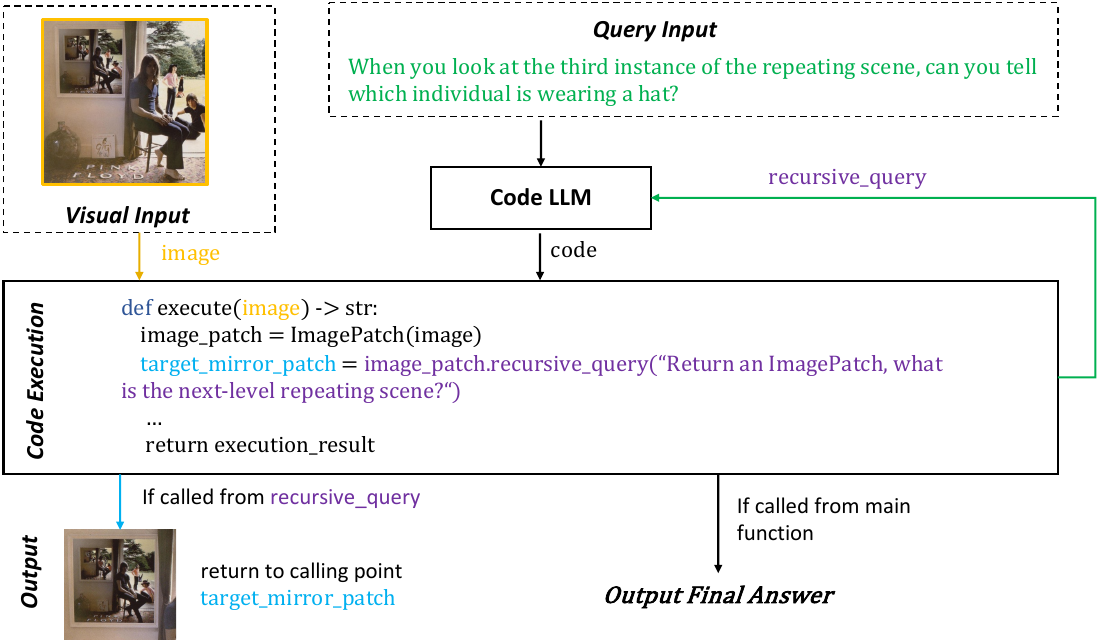}
    \caption{\textbf{An overview of our approach.} When a \texttt{recursive\_query} API call is occured, the model generates and executes new code based on the query, returning the output back to the original call.}
    \label{fig:model_pipeline_general}
\end{figure}

\section{Recursive Visual Programming}
In this section, we introduce our method, Recursive Visual Programming (RVP). We first review the problem setup and ingredients of visual programming as established by prior work (see Section~\ref{sec:expr:background}). We then introduce our recursive coding framework (see Section~\ref{sec:expr:rvp}), which involves a multi-step prompting scheme, as well as a system for dynamic type assignment (see Section~\ref{sec:expr:dynamic}).
\subsection{Preliminaries}
\label{sec:expr:background}
In Visual Question Answering (VQA), the system $f$ is given a visual input $I$ (\ie an image, a set of images, or a video), as well as a question $Q$ relating to the visual content. The expected output is a textual answer $A$ to the question:
\begin{equation}
    A \leftarrow f(I, Q)
\end{equation}
Visual Programming methods \cite{visprog,codevqa,vipergpt} represent $f$ as the composition of two functions: a code generator $g$ and a visual executor $h$. In the first part, the code generator generates a program $P$ conditioned on the question, and in the second part, the executor executes the program on the visual input:
\begin{equation}
    P \leftarrow g(Q), A \leftarrow h(P, I)
\end{equation}
We focus on the few-shot setting in which we have only a few (at most 50) training examples--each consisting of a question, program, and answer--for a given task. Therefore, the code generator $g$ is composed of a language model and a prompt that includes a description of the API available to the executor and in-context examples of questions along with the corresponding correct programs. We use Python as the programming language, similar to \cite{codevqa,vipergpt}.

We evaluate our approach using the ViperGPT \cite{vipergpt} API. For questions over images or sets of images, these functions can be called on an \texttt{ImagePatch} object, which can be any crop of the image:
\begin{enumerate}
\item \texttt{find(object: str)}: Returns a list of \texttt{ImagePatch} objects matching the description.
\item \texttt{exists(object: str)}: Returns a bool representing whether the described object exists.
\item \texttt{verify\_property(object: str, property: str)}: Returns a bool representing whether the described object has the specified property.
\item \texttt{simple\_query(question: str)}: Returns an answer to the provided question based on the given \texttt{ImagePatch}.
\item \texttt{compute\_depth()}: Returns the median depth of the region in the given \texttt{ImagePatch}.
\item \texttt{crop(left: int, lower: int, right: int, upper: int)}: Crops a part of the given \texttt{ImagePatch} and returns a new \texttt{ImagePatch} representing that crop.
\end{enumerate}
For questions about a video, the API includes three other functions: \texttt{trim}(), \texttt{frame\_from\_index}(), and \texttt{frame\_iterator}(), which return an abbreviated version of the video, a specific frame, and an iterator over the frames respectively.
\subsection{Recursive Visual Programming}
\label{sec:expr:rvp}
We introduce a new method for code generation in visual programming: Recursive Visual Programming (RVP). RVP enables the language model to break the code generation process into multiple sub-questions and recursively call itself to generate a program for each sub-question. As shown in Figure~\ref{fig:model_pipeline_general}, we operationalize RVP by adding a new function to the API: \texttt{recursive\_query(\text{image}, \text{sub-question})}. 
During execution, when \texttt{recursive\_query} is called, it prompts the language model to generate a new code given the sub-question. This code is then executed, and its return value is returned by \texttt{recursive\_query}. An example break down of the usage of \texttt{recursive\_query} call is shown in Figure~\ref{fig:model_pipeline_example}. We set the recursive termination condition by (1) setting max depth of 10 and (2) switching to direct simple query when generated code contains a recursive call to the same input question.

To enable this process, we add examples of calling \texttt{recursive\_query} in the prompt to show the model how to decompose the problem and utilize the answers returned by \texttt{recursive\_query}. We provide examples of the in-context prompt of \texttt{recursive\_query} we used in the supplementary material.
\begin{figure}[htbp]
    \centering
    \includegraphics[width=12cm]{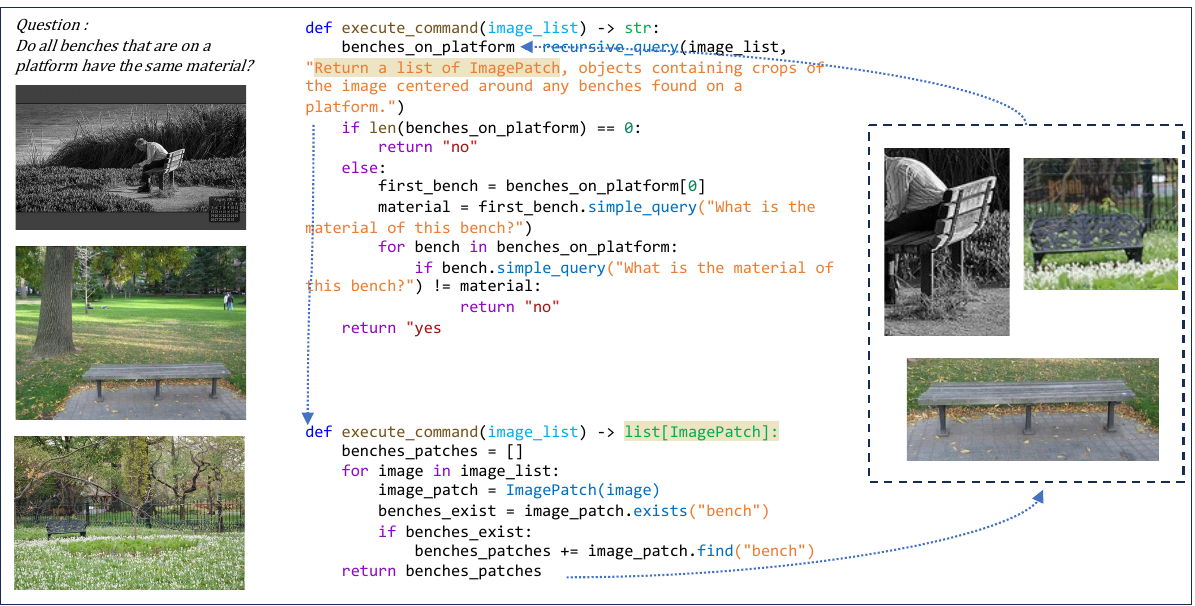}
    \caption{\textbf{Examples of dynamic type assignment in \texttt{recursive\_query} calls}. The model autonomously determines the appropriate return type in its code generation and generalize to new types unseen in in-context examples such as \texttt{List[ImagePatch]}. }
    \label{fig:dynamic_type_example}
\end{figure}
\subsection{Dynamic Type Assignment}
\label{sec:expr:dynamic}
In previous visual programming frameworks, code generation mechanisms are constrained by static type assignments, \ie, the return type of a given function is fixed. For instance, \texttt{simple\_query} always returns a string.
However, this rigid structure limits the model's flexibility to adapt the code to different contexts within the VQA task. RVP introduces dynamic type assignment to address this limitation. When the code generator writes a \texttt{recursive\_query} function, it can include the expected return type as a part of the question that is passed to the \texttt{recursive\_query}. To facilitate this, in the in-context examples that include \texttt{recursive\_query}, we let the input `question' specify a type. As shown in Figure~\ref{fig:dynamic_type_example}, we give examples of type ImagePatch, bool, float, and str being specified within the queries. When receiving this question, the model discerns the required type and generates code that declares and returns the correct type.
For instance, a query like ``Return a bool, is there a black cat in the picture?" will prompt the generation of code with the signature.
\begin{equation}
\text{execute}(\text{image}) \rightarrow \text{bool}
\end{equation}
This signature indicates that the generated code \( P' \), when executed, will return a boolean value \( A_{\text{bool}} \) which answers the sub-question \( Q' \), illustrating the dynamic and context-sensitive nature of RVP's code generation capabilities that help to produce syntactically and semantically appropriate code for complex reasoning.
\section{Experiments} 
In this section, we introduce the basic settings including the task formulation and parameter details. Then, we provide experiment results and in-depth analysis.
\vspace{-10pt}
\subsection{Experiment Setup}
\minisection{Datasets} We evaluate on four datasets spanning a wide range of visual reasoning skills and types of visual input: Visual Spatial Reasoning (VSR) \cite{vsr}, GQA \cite{gqa}, COVR \cite{covr}, and NextQA \cite{nextqa}. For NextQA, we present results on the ``Hard Temporal'' subset which was curated to consist of questions that truly require temporal reasoning \cite{atp}. VSR and GQA include queries about single images, while COVR includes queries about multiple images, and NextQA includes queries about video. On each dataset, we evaluate methods by their accuracy in producing an answer that exactly matches the ground-truth answer. For NextQA, we adopt a multiple-choice setup in which we provide methods with the set of possible answer choices for each question. For VSR, a spatial statement of an image is made, and the question is framed as : ``Is it true or false, '' followed by the statement. In this way, a VSR statement is framed as a VQA question and the model will answer either true or false.

\minisection{Implementation Details} We implement RVP using the ViperGPT API and the accompanying prompt template. For each dataset, we curate a set of in-context examples which are added to the prompt. The set of question-answer pairs is shared between the recursive and non-recursive methods. We use seven in-context examples for GQA, six for NextQA, six for VSR, and eight for COVR. For code generation, use GPT-3.5-turbo via the OpenAI API \footnote{www.platform.openai.com}.
\subsection{Quantitative Results}
Table~\ref{tab:main-results} presents our results across four datasets. In all settings, we outperform ViperGPT, which is identical to RVP except that RVP includes \texttt{recursive\_query} function in the API and in-context examples. This indicates that recursive programming indeed improves accuracy in visual reasoning tasks. Moreover, our method outperforms previous zero/few-shot methods in all but one of the settings.  While our primary aim was not to push the state-of-the-art results, we show that integrating the clarity and elegance of recursive coding into standard VP methods not only doesn't impair performance, but can enhance it.
\begin{table}[htbp]
\scriptsize
\centering
\caption{\textbf{Results.} We report exact-match accuracy on each dataset. Best zero/few-shot result for each dataset is \textbf{bolded}. \textdagger Fully supervised results are from different models: ViLT \cite{vilt} for VSR, HiTeA \cite{hitea} for NextQA, VinVL-Base \cite{vinvl} for GQA, and VisualBERT \cite{visualbert} for COVR. \textdaggerdbl We report the results of our reproduction of ViperGPT using the officially released code and the same in-context examples used for our method (since the original in-context examples from ViperGPT were not released). We use GPT3.5-turbo while CodeVQA uses Codex, which is not available anymore.}
\begin{tabular}{@{}lcccccc@{}}
\toprule
\multirow{2}[3]{*}{{\bf Method}} & \multicolumn{2}{c}{VSR} & \multicolumn{1}{c}{NextQA} & \multicolumn{1}{c}{GQA} & \multicolumn{1}{c}{COVR} \\
 & Random Split & Zero-shot Split & Hard Split-T &  Test-dev & Test \\
\midrule
Fully supervised\textdagger & 69.3 & 63.0 & 48.6  & 65.1 & 57.9 \\
\midrule
\textit{Zero/few-shot methods} & & & & & & \\
CLIP & 56.0 & 54.5 & -- & -- & -- \\
BLIPv2 & -- & -- & -- & 42.31 & -- \\
CodeVQA & -- & -- & --  & \textbf{49.0} & 50.7 \\
ViperGPT\textdaggerdbl & 61.25 & 61.59 & 47.21 & 44.63 & 51.69 \\
RVP (ours) & 63.53 & \textbf{66.09} & \textbf{48.82}  & 45.62 & \textbf{52.67} \\
\bottomrule
\end{tabular}
\label{tab:main-results}
\end{table}
\subsection{Qualitative Results}
\begin{figure}[h]
\centering
\includegraphics[width=\textwidth]
{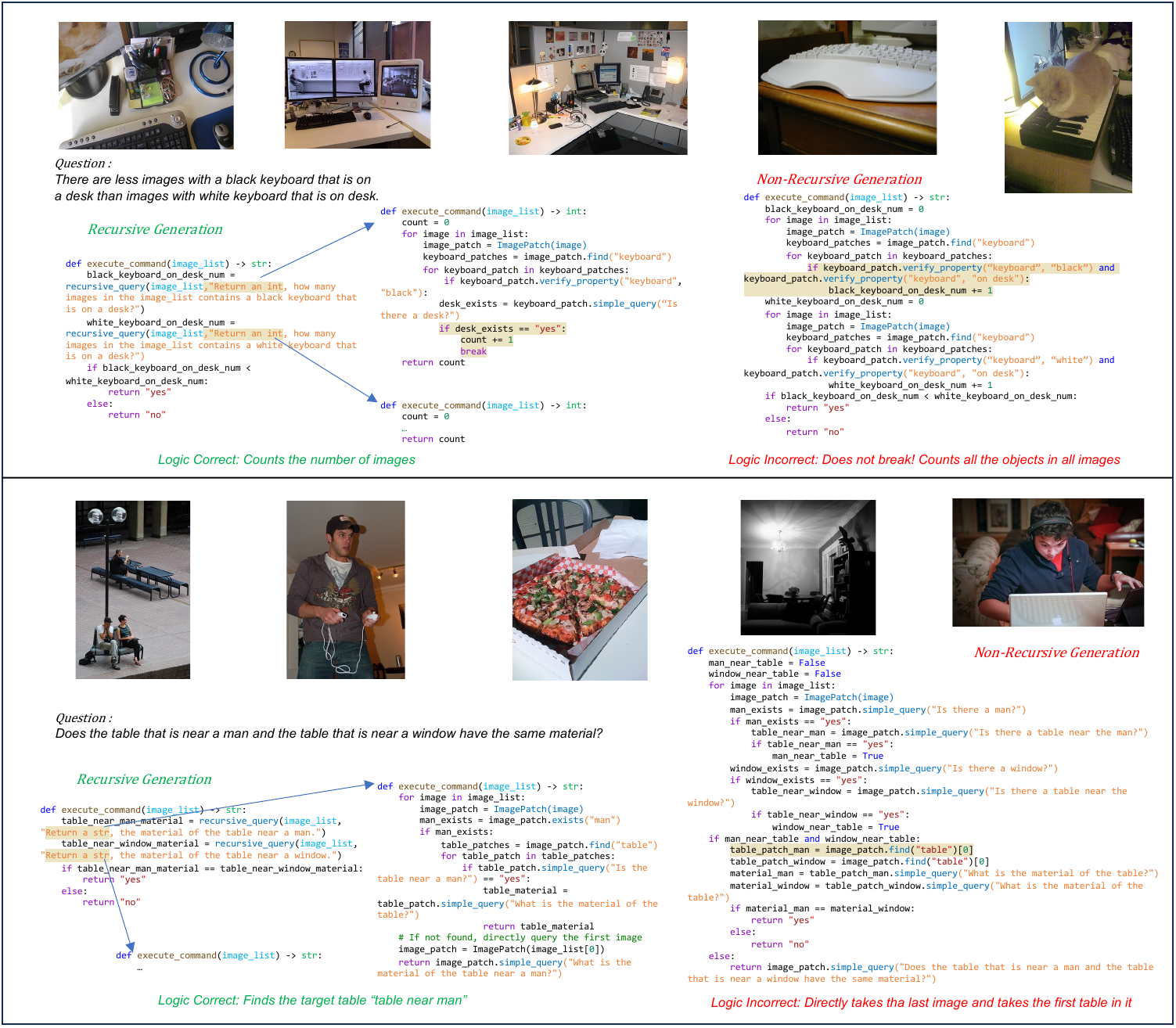}
\caption{\textbf{Examples from COVR.} RVP allows correct logic flow while traditional non-recursive VP fails to handle the logic correctly.}
\label{fig:covr-example}
\vspace{-10pt}
\end{figure}
\begin{figure}[h]
\centering
\includegraphics[width=12cm]
{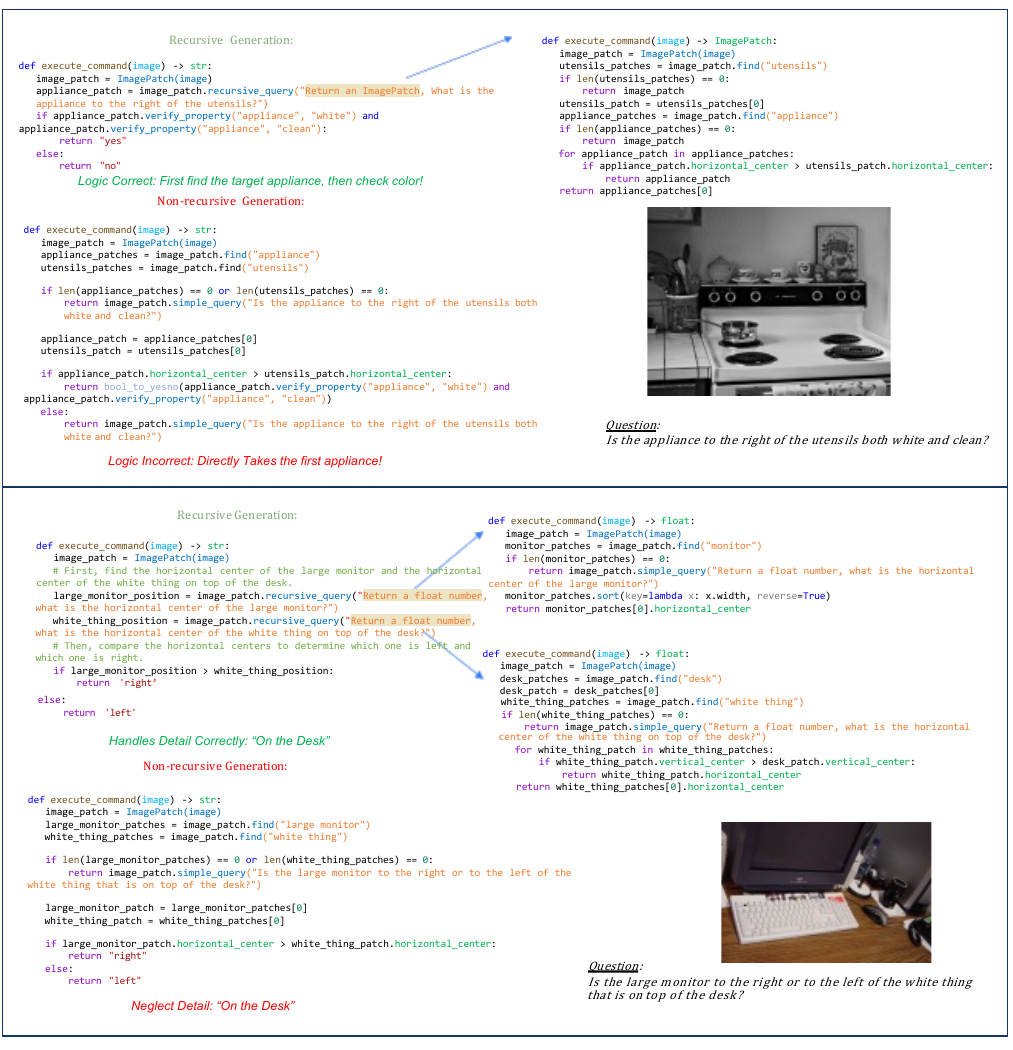}
\caption{\textbf{Examples from GQA.} RVP allows handling the details more elegantly and produces logically correct code.}
\label{fig:gqa-example}
\end{figure}
We provide qualitative examples in Figure~\ref{fig:teaser},\ref{fig:covr-example},\ref{fig:gqa-example}.
In the example from COVR in Figure \ref{fig:teaser}, the non-recursive approach answers with neither ``yes'' nor ``no,'' while the recursive approach does confirm to these choices. One possible explanation for this issue is that LLMs struggle to maintain coherence across long contexts \cite{doc}. RVP circumvents this issue by pushing the LLM to commit to the answer types earlier in its generation (the length from the beginning of the code to the point  ``return xxx'' is shorter). In Figure \ref{fig:covr-example}, the non-recursive approach fails to handle the logic correctly while the recursive approach correctly does. This is likely due to that RVP breaks down the code logic and each code piece only needs to focus on a single logic, while the non-recursive method attempts to handle all the logic at once. In the GQA example, a common pattern involves handling complex logical problems and multiple properties. Non-recursive methods often fell short in these cases, either overlooking critical details or processing them in a logically flawed manner. RVP manages these intricate questions by addressing each detail accurately. This difference may stem from the inherent difficulty LLMs face in processing complicated logic over extended contexts. Recursive methods simplify this by focusing on the overall logic and gradually breaking the problems down into smaller, more manageable segments. This facilitates easier and more accurate resolution.

We provide a dynamic type assignment example in Figure~\ref{fig:dynamic_type_example}. In this example, RVP assigns the return type \texttt{List[ImagePatch]} in the \texttt{recursive\_query} call. It's worth noting that the return type \texttt{List[ImagePatch]} is unseen in the in-context examples. This suggests that RVP has the potential to utilize complicated and diverse data structures, highlight the effectiveness of our approach.
\subsection{Ablation Study}
\paragraph{Dynamic Type} This section studies integrating dynamic type assignment within the \texttt{recursive\_query} API. For the GQA dataset, we examined all instances with recursive patterns in the \texttt{test\_dev} split. In COVR, where the test set is unavailable, we used the \texttt{val\_1000} samples from the validation set, following \cite{codevqa}.

We compare four approaches: \textbf{Non-recursive Baseline}: Standard approach without recursive features. \textbf{Fixed-Type Recursive}: The \texttt{recursive\_query} API is constrained to return a string. \textbf{Dynamic-Implicit Recursive}: The API can return various types, but the specific type is not predefined in the query. For example, ``Is there a black cat?" could yield either a boolean or a string. \textbf{Dynamic-Explicit Recursive}: Our primary approach, where the API's input query specifies the return type, like ``Return a bool, is there a black cat?".

Table \ref{tab:dynamic_ablation} shows our findings. The dynamic-explicit type assignment outperformed other methods, underscoring the need for clear type specification. The dynamic-implicit method showed lower performance than non-recursive due to type ambiguities, as exemplified in Figure \ref{fig:type_error}, where the first code piece assumes the return type is a bool, while the second code piece returns a str. Fixed-type assignment proves to be beneficial over non-recursive methods, indicating that recursion itself adds value, but its rigidity in type handling limits performance.
This study highlights the benefits of flexible, dynamic type assignment and the crucial role of explicit type declaration for accurate and consistent responses. These insights mark a step towards more adaptable visual reasoning frameworks, emphasizing the need for precision in dynamic environments.
\paragraph{In-Context Examples} We conduct ablation on using retrieval-based in-context example selection and using different number of in-context examples. \\
\textbf{Retrieval-based in-context example selection.} We manage to write 9 programs using the recursive method from the 50 provided by CodeVQA \cite{codevqa}, and add our original 3 recursive programs to them. We keep the non-recursive in-context example unchanged and choose 3 recursive in-context examples from these 12 for each question.
We follow \cite{wang2022language} to use the embedding-based retrieval technique that calculates a score for each example, and then select the top 3 examples for each question. 
The results of \texttt{GQA\_val\_2000} are in Table \ref{tab:retrieval}. We find that the recursive pattern occurance rate is significantly higher than using fixed in-context examples. This is likely because by retrieving similar recursively-decomposed questions, the model can more easily simulate their patterns, thereby attempting to solve more problems recursively. However. the overall accuracy does not improve. This suggests that although the retrieval-based method encourages more diverse and suitable patterns of recursive decomposition, there is a need for improved techniques in selecting the example pool—-such as focusing on those most suitable for recursive decomposition instead of randomly selecting—along with careful tuning of the example programs. \\
\textbf{Number of in-context examples.} We randomly add one example/delete one example from the \texttt{covr\_1000\_val}, the results are shown in Table \ref{tab:retrieval}. The results indicate that adding the number of examples can enhance the performance, but does not have a significant impact.
\paragraph{Error feedback loop} Recent studies have studied the debugging ability of LLMs on code\cite{chen2023teaching}. We study whether modular visual code structure enables better bug identification for visual programming.
We sample 20 incorrect programs for 10 questions that have type/logic errors, and use GPT3.5 to (1) identify the bug, and (2) write the correct code given the bug. 
For identification, the model correctly finds 7 bugs correctly for RVP and 2 for non-recursive visual programming. For correction, the model fixes 7 bugs correctly for RVP and 7 for visual programming. This suggests that clearer and more modular code structure, as presented RVP, can potentially aids bug identification.
\begin{table}[htbp]
\vspace{-10pt}
\centering
\small
\caption{\textbf{Ablation study.} (a) We provide ablation study results for dynamic type assignment and found that specified dynamic type assignment in RVP achieves the highest performance. (b) We provide results on using retrieval-based in-context examples and fixed examples. Retrieval-based method enables more diverse recursive patterns but requires careful choosing the prompt pool. (c) We provide results of using different number of examples and found this doesn't influence performance significantly.}
\begin{minipage}{.4\linewidth}
\centering
\caption*{(a) Dynamic Type Ablation.}
\begin{tabular}{@{}lcc@{}}
\toprule
Method & GQA & COVR  \\
\midrule
Non-Recursive & 55.99 & 60.71\\
Fixed-Type & 65.05 & 60.12\\
Dynamic-Implicit & 55.34 & 53.57\\
Dynamic-Explicit & \textbf{70.23} & \textbf{67.86}\\
\bottomrule
\label{tab:dynamic_ablation}
\end{tabular}
\end{minipage}%
\hfill 
\begin{minipage}{.3\linewidth}
\centering
\caption*{(b) Retrieval-Based and Fixed Examples}
\begin{tabular}{@{}lcc@{}}
\toprule
 & Fixed & Retrieval\\
\midrule
Accuracy & 54.75 & 53.79\\
Rec Rate & 3.55\% & 14.85\%\\
Rec Acc & 71.83 & 57.91\\
\bottomrule
\label{tab:retrieval}
\end{tabular}
\end{minipage}%
\hfill 
\begin{minipage}{.25\linewidth}
\centering
\caption*{(c) In-Context Example Num}
\begin{tabular}{@{}lc@{}}
\toprule
Method & Accuracy \\
\midrule
Del 1 & 50.65\\ 
Original & 51.05\\
Add 1 & 51.35\\
\bottomrule
\label{tab:retrieval}
\end{tabular}
\end{minipage}%
\vspace{-30pt}
\end{table}
\subsection{Readability Study}
This study investigates how recursive coding impacts code readability and comprehension for human programmers. Our goal is to move from complex code structures to more modular, clear, and understandable code. To evaluate this, we focus on code quality in terms of its understandability.
\begin{figure}[htbp]
\centering
\begin{minipage}[t]{.48\textwidth} 
  \centering
  \includegraphics[width=\linewidth]{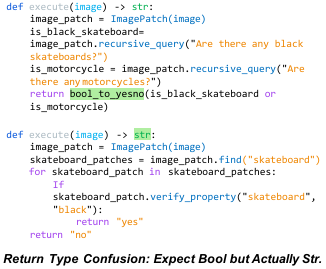} 
  \captionof{figure}{\textbf{Failure example of dynamic type assignment} due to unspecified type, The model confuses the type and return a str instead of a bool.}
  \label{fig:type_error}
\end{minipage}%
\hfill 
\begin{minipage}[t]{.5\textwidth} 
  \centering
  \includegraphics[width=\linewidth]{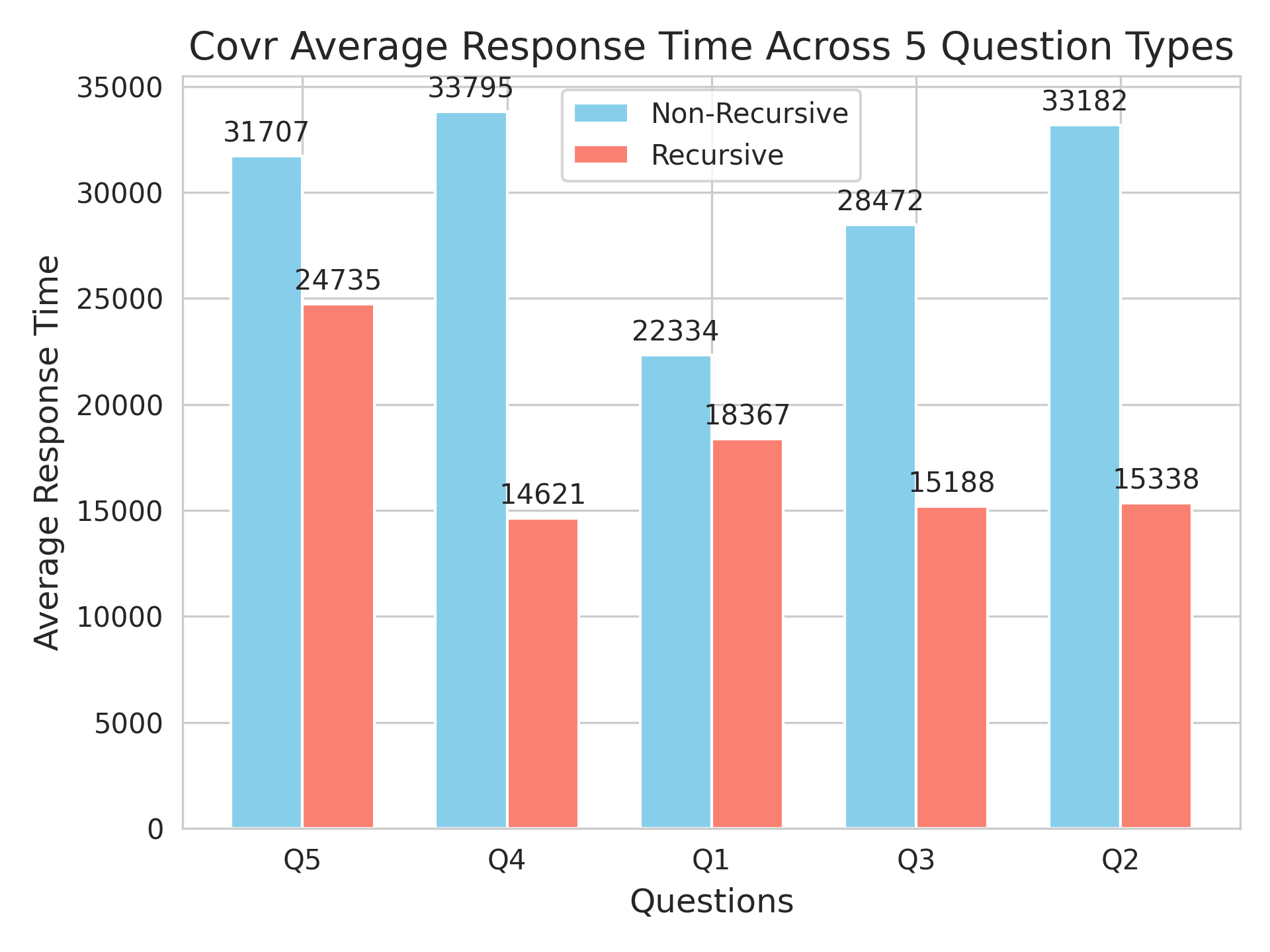} 
  \captionof{figure}{Comparison of average understanding time for recursive and non-recursive codes. RVP demonstrates better readability and is easy to understand.}
  \label{fig:survey}
\end{minipage}
\end{figure}

From the COVR dataset, we randomly select one example from each five different question types where recursive patterns occur. We present the main function of both their recursive and non-recursive code versions to 15 participants and ask them to grasp the high-level intent of the code as quickly as possible. For the non-recursive case, the main function is just the single code snippet generated by the LLM. For the recursive case, the main function consists of the code generated by the first LLM call (the high level code) but not the functions generated by the recursive queries. This helps us measure how fast users can understand the code's intention by looking at the main function, and reflects the self-explanatory nature of the code. Figure \ref{fig:survey} shows the results. Participants consistently understand the recursive code faster than the non-recursive versions. This quicker comprehension can be attributed to the recursive code's structured and clear approach. Non-recursive code, while correct, sometimes lacks immediate clarity in conveying its primary goal.

We also calculate the average line length for all main and sub functions. For RVP, the average length of all functions is \textbf{8.47}. For non-recursive VP, the average length of the functions is \textbf{15.4}. This suggests that RVP decreases the average length the code pieces and enhances modularity.

Our findings suggest that recursive coding not only simplifies complex code but also enhances its accessibility and ease of understanding. Its orderly structure and clear logic accelerate comprehension, leading to more efficient and user-friendly programming. We provide more details of this study in the supp.

\subsection{Recursive Analysis}
\begin{table}[htbp]
\centering
\caption{\textbf{Recursive analysis.} Table (a) provides the distribution of recursive patterns by question type in GQA and COVR. We have found that recursive patterns occur commonly in question types that require reasoning. Table (b) studies different recursive query return types and their frequency in GQA and COVR. The model generalizes to unseen types. Table (c) compares accuracy across question types. Recursive methods outperform non-recursive particularly in ’Choose’ and ’Query’ questions.}
\vspace{-3mm}
\label{tab:questiontype}
\begin{minipage}[t]{.4\linewidth}
\centering
\scriptsize
\caption*{(a) Dist. of Recursive Patterns}
\begin{tabular}{@{}lcc@{}}
    \toprule
    \textbf{COVR/GQA} & \multicolumn{2}{c}{\textbf{Question Types}} \\
    \cmidrule(r){2-3}
    & COVR & GQA \\
    \midrule
    Compare & 457 & 15 \\
    Mult. Ref & 282 & - \\
    Quant. Attr & 12 & - \\
    Spec Attr & 357 & - \\
    Choose & 1 & 25 \\
    Verify & - & 68 \\
    Query & - & 167 \\
    Logical & - & 34 \\
    \bottomrule
    \end{tabular}
\end{minipage}%
\hfill 
\begin{minipage}[t]{.3\linewidth}
\centering
\scriptsize
    \caption*{(b) Diverse Return Types}
    \begin{tabular}{@{}lcc@{}}
    \toprule
    \textbf{Data Type} & \textbf{GQA} & \textbf{COVR} \\
    \midrule
    Str & 319 & 1828 \\
    Bool & 521 & 592 \\
    List[str] & 16 & - \\
    ImagePatch & 30 & - \\
    List[ImgPatch] & 1 & 5 \\
    Float & 6 & - \\
    Int & - & 914 \\
    \bottomrule
    \end{tabular}
\end{minipage}%
\hfill 
\begin{minipage}[t]{.3\linewidth}
\centering
\scriptsize
\caption*{(c) Type Accuracy}
\begin{tabular}{@{}lcc@{}}
\toprule
\textbf{Question} & \multicolumn{2}{c}{\textbf{Accuracy (\%)}} \\
\cmidrule(r){2-3}
& Non-Rec & Rec \\
\midrule
Verify & 52.94 & 64.71 \\
Query & 56.89 & 71.86 \\
Logical & 55.88 & 70.59 \\
Compare & 66.67 & 53.33 \\
Choose & 52.00 & 84.00 \\
\bottomrule
\end{tabular}
\end{minipage}
\vspace{-5mm}
\end{table}
\minisection{Question Type} We study which types of questions tend to use recursive and found that recursive patterns appeared in five question types within both the COVR test set and the GQA test-dev set, as shown in Table \ref{tab:questiontype}.
The 'Verify Property' and 'Query' types in GQA, and 'Compare' and 'Multi-Reference' in COVR, were most common. This suggests recursive methods tend to occur in tasks that require comparative and referential analysis.\\
\minisection{Dynamic Type} We study which types of dynamic return types occurs and their frequently in GQA and COVR. In Table \ref{tab:questiontype}, we found that RVP generalizes to new types like \texttt{List[str]} and \texttt{List[ImagePatch]}, which are not included in the in-context examples. This flexibility and ability to generalize beyond in-context examples highlight the inherent adaptability of recursive methods.\\
\minisection{Type Accuracy} We evaluate the accuracy of recursive methods across different question types. An analysis of the GQA test-dev set is presented in Table \ref{tab:questiontype}. It reveals that recursive methods significantly improve accuracy in 'Query', 'Logical', and 'Choose' questions. This underscores their suitability for complex reasoning and decision-making tasks.\\
\minisection{Error Source} We sample 50 incorrect recursive examples from \texttt{GQA2000val} and manually checke the error source: 22\% are caused by logic error inside sub-functions; 18\% are caused by logic error between functions; 52\% are caused by API model inaccuracy (i.e. detection model error); and 4\% are caused by other problems like the groundtruth answer unclear. \\
\minisection{Cost And Runtime} For the questions which require recursive query, RVP takes about 2.1 times (runtime) and about 3 times (cost) compared to ViperGPT.
\vspace{-15pt}
\section{Conclusion and Limitation} 
In this work, we present RVP, a visual programming method that adopts recursive coding to generate concise, elegant and easy-to-understand code,  handling details and intricate logic more accurately compared with traditional visual programming methods. We conduct extensive experiments across various benchmarks and provide a comprehensive analysis of RVP. Our work is the first to explore recursive programming in the visual domain and dynamic-type code generation for visual reasoning; we believe the recursive programming concept would have a potential greater impact on other domains as well. While RVP is effective for complicated visual questions, most questions in current VQA benchmarks tend to be short and non-recursive. We believe that the benefit of our method could be demonstrated better with a new benchmark or a more complex split.


\section*{Acknowledgements}
We would like to thank Ben Bogin for helping us test on the private COVR test set, and Guangyuan Jiang for helping us proof read the paper and providing valuable feedback. 
\bibliographystyle{splncs04}
\bibliography{main}

\begin{thebibliography}{10}
\providecommand{\url}[1]{\texttt{#1}}
\providecommand{\urlprefix}{URL }
\providecommand{\doi}[1]{https://doi.org/#1}

\bibitem{nmn}
Andreas, J., Rohrbach, M., Darrell, T., Klein, D.: Neural module networks. 2016 IEEE Conference on Computer Vision and Pattern Recognition (CVPR) pp. 39--48 (2015), \url{https://api.semanticscholar.org/CorpusID:5276660}

\bibitem{andreas2016}
Andreas, J., Rohrbach, M., Darrell, T., Klein, D.: Learning to compose neural networks for question answering. ArXiv  \textbf{abs/1601.01705} (2016), \url{https://api.semanticscholar.org/CorpusID:3130692}

\bibitem{antol2015vqa}
Antol, S., Agrawal, A., Lu, J., Mitchell, M., Batra, D., Zitnick, C.L., Parikh, D.: Vqa: Visual question answering. In: Proceedings of the IEEE international conference on computer vision. pp. 2425--2433 (2015)

\bibitem{besta2024graph}
Besta, M., Blach, N., Kubicek, A., Gerstenberger, R., Podstawski, M., Gianinazzi, L., Gajda, J., Lehmann, T., Niewiadomski, H., Nyczyk, P., Hoefler, T.: Graph of thoughts: Solving elaborate problems with large language models (2024)

\bibitem{covr}
Bogin, B., Gupta, S., Gardner, M., Berant, J.: Covr: A test-bed for visually grounded compositional generalization with real images. ArXiv  \textbf{abs/2109.10613} (2021), \url{https://api.semanticscholar.org/CorpusID:237592834}

\bibitem{neurips2020gpt3}
Brown, T., Mann, B., Ryder, N., Subbiah, M., Kaplan, J.D., Dhariwal, P., Neelakantan, A., Shyam, P., Sastry, G., Askell, A., Agarwal, S., Herbert-Voss, A., Krueger, G., Henighan, T., Child, R., Ramesh, A., Ziegler, D., Wu, J., Winter, C., Hesse, C., Chen, M., Sigler, E., Litwin, M., Gray, S., Chess, B., Clark, J., Berner, C., McCandlish, S., Radford, A., Sutskever, I., Amodei, D.: Language models are few-shot learners. In: Larochelle, H., Ranzato, M., Hadsell, R., Balcan, M., Lin, H. (eds.) Advances in Neural Information Processing Systems. vol.~33, pp. 1877--1901. Curran Associates, Inc. (2020), \url{https://proceedings.neurips.cc/paper/2020/file/1457c0d6bfcb4967418bfb8ac142f64a-Paper.pdf}

\bibitem{atp}
Buch, S., Eyzaguirre, C., Gaidon, A., Wu, J., Fei-Fei, L., Niebles, J.C.: {Revisiting the ``Video'' in Video-Language Understanding}. In: Proceedings of the IEEE/CVF Conference on Computer Vision and Pattern Recognition (CVPR) (2022)

\bibitem{codex}
Chen, M., Tworek, J., Jun, H., Yuan, Q., Ponde, H., Kaplan, J., Edwards, H., Burda, Y., Joseph, N., Brockman, G., Ray, A., Puri, R., Krueger, G., Petrov, M., Khlaaf, H., Sastry, G., Mishkin, P., Chan, B., Gray, S., Ryder, N., Pavlov, M., Power, A., Kaiser, L., Bavarian, M., Winter, C., Tillet, P., Such, F.P., Cummings, D.W., Plappert, M., Chantzis, F., Barnes, E., Herbert-Voss, A., Guss, W.H., Nichol, A., Babuschkin, I., Balaji, S.A., Jain, S., Carr, A., Leike, J., Achiam, J., Misra, V., Morikawa, E., Radford, A., Knight, M.M., Brundage, M., Murati, M., Mayer, K., Welinder, P., McGrew, B., Amodei, D., McCandlish, S., Sutskever, I., Zaremba, W.: Evaluating large language models trained on code. ArXiv  \textbf{abs/2107.03374} (2021), \url{https://api.semanticscholar.org/CorpusID:235755472}

\bibitem{chen2022Program}
Chen, W., Ma, X., Wang, X., Cohen, W.W.: Program of thoughts prompting: Disentangling computation from reasoning for numerical reasoning tasks. Transactions on Machine Learning Research  (2023)

\bibitem{chen2023teaching}
Chen, X., Lin, M., Sch{\"a}rli, N., Zhou, D.: Teaching large language models to self-debug. arXiv preprint arXiv:2304.05128  (2023)

\bibitem{binder}
Cheng, Z., Xie, T., Shi, P., Li, C., Nadkarni, R., Hu, Y., Xiong, C., Radev, D.R., Ostendorf, M., Zettlemoyer, L., Smith, N.A., Yu, T.: Binding language models in symbolic languages. ArXiv  \textbf{abs/2210.02875} (2022), \url{https://api.semanticscholar.org/CorpusID:252734772}

\bibitem{vpgen}
Cho, J., Zala, A., Bansal, M.: Visual programming for text-to-image generation and evaluation. NeurIPS  (2023)

\bibitem{gao2022Pal}
Gao, L., Madaan, A., Zhou, S., Alon, U., Liu, P., Yang, Y., Callan, J., Neubig, G.: Pal: Program-aided language models. arXiv preprint arXiv:2211.10435  (2022)

\bibitem{visprog}
Gupta, T., Kembhavi, A.: Visual programming: Compositional visual reasoning without training. 2023 IEEE/CVF Conference on Computer Vision and Pattern Recognition (CVPR) pp. 14953--14962 (2022), \url{https://api.semanticscholar.org/CorpusID:253734854}

\bibitem{hu2018}
Hu, R., Andreas, J., Rohrbach, M., Darrell, T., Saenko, K.: Learning to reason: End-to-end module networks for visual question answering. 2017 IEEE International Conference on Computer Vision (ICCV) pp. 804--813 (2017), \url{https://api.semanticscholar.org/CorpusID:18682}

\bibitem{Hudson2019GQAAN}
Hudson, D.A., Manning, C.D.: Gqa: A new dataset for real-world visual reasoning and compositional question answering. 2019 IEEE/CVF Conference on Computer Vision and Pattern Recognition (CVPR) pp. 6693--6702 (2019), \url{https://api.semanticscholar.org/CorpusID:152282269}

\bibitem{gqa}
Hudson, D.A., Manning, C.D.: Gqa: A new dataset for real-world visual reasoning and compositional question answering. 2019 IEEE/CVF Conference on Computer Vision and Pattern Recognition (CVPR) pp. 6693--6702 (2019), \url{https://api.semanticscholar.org/CorpusID:152282269}

\bibitem{hudson2019}
Hudson, D.A., Manning, C.D.: Learning by abstraction: The neural state machine. In: Neural Information Processing Systems (2019), \url{https://api.semanticscholar.org/CorpusID:195847902}

\bibitem{vilt}
Kim, W., Son, B., Kim, I.: Vilt: Vision-and-language transformer without convolution or region supervision. In: International Conference on Machine Learning (2021), \url{https://api.semanticscholar.org/CorpusID:231839613}

\bibitem{visualbert}
Li, L.H., Yatskar, M., Yin, D., Hsieh, C.J., Chang, K.W.: Visualbert: A simple and performant baseline for vision and language. ArXiv  \textbf{abs/1908.03557} (2019), \url{https://api.semanticscholar.org/CorpusID:199528533}

\bibitem{codeAsPolicies}
Liang, J., Huang, W., Xia, F., Xu, P., Hausman, K., Ichter, B., Florence, P., Zeng, A.: Code as policies: Language model programs for embodied control. In: arXiv preprint arXiv:2209.07753 (2022)

\bibitem{vsr}
Liu, F., Emerson, G.E.T., Collier, N.: Visual spatial reasoning. Transactions of the Association for Computational Linguistics  \textbf{11},  635--651 (2022), \url{https://api.semanticscholar.org/CorpusID:248496506}

\bibitem{liu2023lost}
Liu, N.F., Lin, K., Hewitt, J., Paranjape, A., Bevilacqua, M., Petroni, F., Liang, P.: Lost in the middle: How language models use long contexts (2023)

\bibitem{chameleon}
Lu, P., Peng, B., Cheng, H., Galley, M., Chang, K.W., Wu, Y.N., Zhu, S.C., Gao, J.: Chameleon: Plug-and-play compositional reasoning with large language models. arXiv preprint arXiv:2304.09842  (2023)

\bibitem{eureka}
Ma, Y.J., Liang, W., Wang, G., Huang, D.A., Bastani, O., Jayaraman, D., Zhu, Y., Fan, L., Anandkumar, A.: Eureka: Human-level reward design via coding large language models. arXiv preprint arXiv: Arxiv-2310.12931  (2023)

\bibitem{madaan2022Language}
Madaan, A., Zhou, S., Alon, U., Yang, Y., Neubig, G.: Language models of code are few-shot commonsense learners. In: Goldberg, Y., Kozareva, Z., Zhang, Y. (eds.) Proceedings of the 2022 Conference on Empirical Methods in Natural Language Processing. pp. 1384--1403. Association for Computational Linguistics, Abu Dhabi, United Arab Emirates (Dec 2022). \doi{10.18653/v1/2022.emnlp-main.90}, \url{https://aclanthology.org/2022.emnlp-main.90}

\bibitem{Marino2019OKVQAAV}
Marino, K., Rastegari, M., Farhadi, A., Mottaghi, R.: Ok-vqa: A visual question answering benchmark requiring external knowledge. 2019 IEEE/CVF Conference on Computer Vision and Pattern Recognition (CVPR) pp. 3190--3199 (2019), \url{https://api.semanticscholar.org/CorpusID:173991173}

\bibitem{OpenAI2023GPT4TR}
OpenAI: Gpt-4 technical report. ArXiv  \textbf{abs/2303.08774} (2023), \url{https://api.semanticscholar.org/CorpusID:257532815}

\bibitem{selfask}
Press, O., Zhang, M., Min, S., Schmidt, L., Smith, N.A., Lewis, M.: Measuring and narrowing the compositionality gap in language models. ArXiv  \textbf{abs/2210.03350} (2022), \url{https://api.semanticscholar.org/CorpusID:252762102}

\bibitem{rozière2023code}
Rozière, B., Gehring, J., Gloeckle, F., Sootla, S., Gat, I., Tan, X.E., Adi, Y., Liu, J., Remez, T., Rapin, J., Kozhevnikov, A., Evtimov, I., Bitton, J., Bhatt, M., Ferrer, C.C., Grattafiori, A., Xiong, W., Défossez, A., Copet, J., Azhar, F., Touvron, H., Martin, L., Usunier, N., Scialom, T., Synnaeve, G.: Code llama: Open foundation models for code (2023)

\bibitem{obtaining2020}
Subramanian, S., Bogin, B., Gupta, N., Wolfson, T., Singh, S., Berant, J., Gardner, M.: Obtaining faithful interpretations from compositional neural networks. In: Annual Meeting of the Association for Computational Linguistics (2020), \url{https://api.semanticscholar.org/CorpusID:218487535}

\bibitem{codevqa}
Subramanian, S., Narasimhan, M., Khangaonkar, K., Yang, K., Nagrani, A., Schmid, C., Zeng, A., Darrell, T., Klein, D.: Modular visual question answering via code generation. In: Rogers, A., Boyd-Graber, J., Okazaki, N. (eds.) Proceedings of the 61st Annual Meeting of the Association for Computational Linguistics (Volume 2: Short Papers). pp. 747--761. Association for Computational Linguistics, Toronto, Canada (Jul 2023). \doi{10.18653/v1/2023.acl-short.65}, \url{https://aclanthology.org/2023.acl-short.65}

\bibitem{vipergpt}
Sur'is, D., Menon, S., Vondrick, C.: Vipergpt: Visual inference via python execution for reasoning. ArXiv  \textbf{abs/2303.08128} (2023), \url{https://api.semanticscholar.org/CorpusID:257505358}

\bibitem{suzgun2022challenging}
Suzgun, M., Scales, N., Sch{\"a}rli, N., Gehrmann, S., Tay, Y., Chung, H.W., Chowdhery, A., Le, Q.V., Chi, E.H., Zhou, D., et~al.: Challenging big-bench tasks and whether chain-of-thought can solve them. arXiv preprint arXiv:2210.09261  (2022)

\bibitem{Touvron2023LLaMAOA}
Touvron, H., Lavril, T., Izacard, G., Martinet, X., Lachaux, M.A., Lacroix, T., Rozi{\`e}re, B., Goyal, N., Hambro, E., Azhar, F., Rodriguez, A., Joulin, A., Grave, E., Lample, G.: Llama: Open and efficient foundation language models. ArXiv  \textbf{abs/2302.13971} (2023), \url{https://api.semanticscholar.org/CorpusID:257219404}

\bibitem{touvron2023llama}
Touvron, H., Martin, L., Stone, K., Albert, P., Almahairi, A., Babaei, Y., Bashlykov, N., Batra, S., Bhargava, P., Bhosale, S., Bikel, D., Blecher, L., Ferrer, C.C., Chen, M., Cucurull, G., Esiobu, D., Fernandes, J., Fu, J., Fu, W., Fuller, B., Gao, C., Goswami, V., Goyal, N., Hartshorn, A., Hosseini, S., Hou, R., Inan, H., Kardas, M., Kerkez, V., Khabsa, M., Kloumann, I., Korenev, A., Koura, P.S., Lachaux, M.A., Lavril, T., Lee, J., Liskovich, D., Lu, Y., Mao, Y., Martinet, X., Mihaylov, T., Mishra, P., Molybog, I., Nie, Y., Poulton, A., Reizenstein, J., Rungta, R., Saladi, K., Schelten, A., Silva, R., Smith, E.M., Subramanian, R., Tan, X.E., Tang, B., Taylor, R., Williams, A., Kuan, J.X., Xu, P., Yan, Z., Zarov, I., Zhang, Y., Fan, A., Kambadur, M., Narang, S., Rodriguez, A., Stojnic, R., Edunov, S., Scialom, T.: Llama 2: Open foundation and fine-tuned chat models (2023)

\bibitem{code4struct}
Wang, X., Li, S., Ji, H.: Code4struct: Code generation for few-shot structured prediction from natural language. arXiv preprint arXiv:2210.12810  (2022)

\bibitem{wang2022language}
Wang, Z., Li, M., Xu, R., Zhou, L., Lei, J., Lin, X., Wang, S., Yang, Z., Zhu, C., Hoiem, D., et~al.: Language models with image descriptors are strong few-shot video-language learners. Advances in Neural Information Processing Systems  \textbf{35},  8483--8497 (2022)

\bibitem{wei2022chain}
Wei, J., Wang, X., Schuurmans, D., Bosma, M., Xia, F., Chi, E., Le, Q.V., Zhou, D., et~al.: Chain-of-thought prompting elicits reasoning in large language models. Advances in Neural Information Processing Systems  \textbf{35},  24824--24837 (2022)

\bibitem{nextqa}
Xiao, J., Shang, X., Yao, A., Chua, T.S.: Next-qa: Next phase of question-answering to explaining temporal actions. In: Proceedings of the IEEE/CVF Conference on Computer Vision and Pattern Recognition (CVPR). pp. 9777--9786 (June 2021)

\bibitem{doc}
Yang, K., Klein, D., Peng, N., Tian, Y.: Doc: Improving long story coherence with detailed outline control. In: Annual Meeting of the Association for Computational Linguistics (2023), \url{https://api.semanticscholar.org/CorpusID:254877751}

\bibitem{yao2023tree}
Yao, S., Yu, D., Zhao, J., Shafran, I., Griffiths, T.L., Cao, Y., Narasimhan, K.: Tree of thoughts: Deliberate problem solving with large language models. arXiv preprint arXiv:2305.10601  (2023)

\bibitem{ye2023comprehensive}
Ye, J., Chen, X., Xu, N., Zu, C., Shao, Z., Liu, S., Cui, Y., Zhou, Z., Gong, C., Shen, Y., Zhou, J., Chen, S., Gui, T., Zhang, Q., Huang, X.: A comprehensive capability analysis of gpt-3 and gpt-3.5 series models (2023)

\bibitem{hitea}
Ye, Q., Xu, G., Yan, M., Xu, H., Qian, Q., Zhang, J., Huang, F.: Hitea: Hierarchical temporal-aware video-language pre-training. ArXiv  \textbf{abs/2212.14546} (2022), \url{https://api.semanticscholar.org/CorpusID:255340506}

\bibitem{languageToRewards}
Yu, W., Gileadi, N., Fu, C., Kirmani, S., Lee, K.H., Gonzalez~Arenas, M., Lewis~Chiang, H.T., Erez, T., Hasenclever, L., Humplik, J., Ichter, B., Xiao, T., Xu, P., Zeng, A., Zhang, T., Heess, N., Sadigh, D., Tan, J., Tassa, Y., Xia, F.: Language to rewards for robotic skill synthesis. Arxiv preprint arXiv:2306.08647  (2023)

\bibitem{vinvl}
Zhang, P., Li, X., Hu, X., Yang, J., Zhang, L., Wang, L., Choi, Y., Gao, J.: Vinvl: Revisiting visual representations in vision-language models. 2021 IEEE/CVF Conference on Computer Vision and Pattern Recognition (CVPR) pp. 5575--5584 (2021), \url{https://api.semanticscholar.org/CorpusID:235692795}

\bibitem{leasttomost}
Zhou, D., Scharli, N., Hou, L., Wei, J., Scales, N., Wang, X., Schuurmans, D., Bousquet, O., Le, Q., hsin Chi, E.H.: Least-to-most prompting enables complex reasoning in large language models. ArXiv  \textbf{abs/2205.10625} (2022), \url{https://api.semanticscholar.org/CorpusID:248986239}

\end{thebibliography}

\clearpage
\title{Supplementary Material: Recursive Visual Programming} 
\author{Jiaxin Ge\inst{1} \quad
Sanjay Subramanian\inst{1} \quad
Baifeng Shi\inst{1}\\ \quad Roei Herzig\inst{1} \quad Trevor Darrell\inst{1}}

\authorrunning{Ge et al.}
\institute{UC Berkeley, CA, USA
}
\maketitle

\section{Overview}
\begin{itemize}
    \item Additional Experiments and Analysis

    \begin{itemize}
        \item Open-sourced Model Performance Study
        \item Error Rate Analysis
        \item Question Complexity Analysis
        \item Prompting Methods Study
        \item In-Context Example Choice Study
    \end{itemize}
    \item Additional Visualization Results
    \item Additional Implementation Details

    \begin{itemize}
        \item Dataset Details
        \item Model Details
        \item Survey Details
        \item In-Context Example Details
    \end{itemize}
\end{itemize}
\section{Additional Experiments and Analysis}
\begin{figure*}
    \centering
\includegraphics[width=12cm]{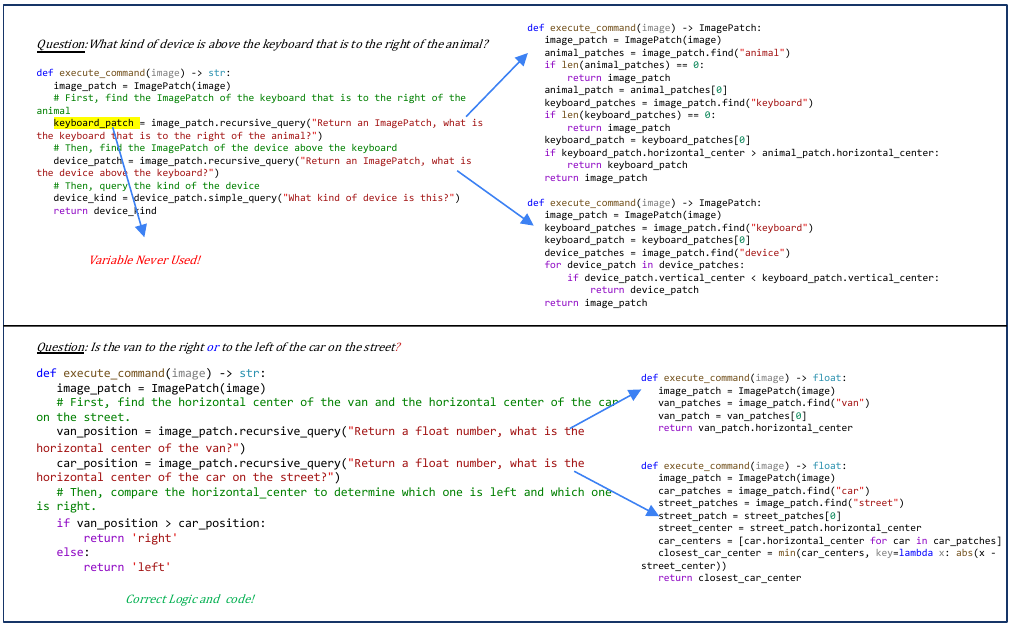}
    \caption{\textbf{Examples of CodeLlama.} In this figure, we provide example codes generated by CodeLlama. While CodeLlama demonstrates impressive reasoning abilities (Below), it also makes certain errors, such as defining unused variables (Above). Interestingly, CodeLlama learns to include logical comments within the code.}
    \label{fig:codellama}
\end{figure*}
\paragraph{Open-sourced Model Performance Study} Previous methods \cite{vipergpt, codevqa} are based on the OpenAI's \texttt{code-davinci-002} model \cite{codex}, which is now unavailable, or \texttt{text-davinci-002} \cite{visprog}. In our main experiments, we use \texttt{gpt-3.5-turbo} model as the code generator. However, this model is not open-sourced. To facilitate further open-sourced studies, we implement our method on one of the most effective open-sourced code generation models, CodeLlama \cite{rozière2023code}, using the same in-context examples.
We test on the GQA val\_2000 sample consistent with CodeVQA. The results are presented in Table ~\ref{tab:model_acc}. It shows that \texttt{gpt-3.5-turbo} model still outperforms the CodeLlama model in terms of accuracy. Furthermore, CodeLlama fails to generalize to unseen types such as List[str] or List[ImagePatch], as shown in Table~\ref{tab:model_type}.
We provide qualitative results of the code generated by CodeLlama in Figure~\ref{fig:codellama}. Generally, CodeLlama demonstrates the ability to code recursively and shows promising reasoning abilities, as shown in the second example of Figure~\ref{fig:codellama}. However, it falls short in coding correctness, such as generating unused variables, as shown in the first example of Figure~\ref{fig:codellama}. Interestingly, we find that CodeLlama consistently writes logical comments within the code for for examples that call recursive\_query, a feature almost never observed in the \texttt{gpt-3.5-turbo} model. This indicates that CodeLlama may possess remarkable in-context learning abilities, but harnessing it might require further prompt tuning of the in-context examples. In our study, we directly use the codes intended for the \texttt{gpt-3.5-turbo} model and only make minor formatting changes for CodeLlama compatibility. To fully explore the ability of CodeLlama, additional format tuning, example selection, or tailored coding may be necessary. We believe these adjustments could help improve the performance of open-sourced models. We leave these to future work.
\begin{table}[htbp]
\small
\centering
\caption{\textbf{Model Accuracy Comparison}: We implement the CodeLlama model and used the same in-context examples to test on \texttt{gpt-3.5-turbo} and CodeLlama. We find that \texttt{gpt-3.5-turbo} attains a higher accuracy than CodeLlama. We also report the scores from CodeVQA which is based on the \texttt{text-davinci-003} model}
\begin{tabular}{@{}lc@{}}
\toprule
 & GQA\_val\_sample \\
\midrule
CodeVQA & 52.5\\
Viper-CodeLlama &  49.12 \\
RVP-CodeLlama& 49.27 \\
Viper-\texttt{gpt-3.5-turbo} & 53.74 \\
RVP-\texttt{gpt-3.5-turbo} & \textbf{54.75} \\
\bottomrule
\end{tabular}
\label{tab:model_acc}
\end{table}
\begin{table}[htbp]
\small
\centering
\caption{\textbf{Diverse return types of CodeLlama and \texttt{gpt-3.5-turbo}.} CodeLlama does not generalize to unseen types like List[str] or List[ImagePatch].}
\begin{tabular}{@{}lcc@{}}
\toprule
Data Type & CodeLlama & \texttt{gpt-3.5-turbo} \\
\midrule
Bool & 101 & 123 \\
List[str] & 0 & 6 \\
ImagePatch & 71 & 1 \\
List[ImagePatch] & 0 & 1\\
Float & 52 & 2 \\
\bottomrule
\end{tabular}
\label{tab:model_type}
\vspace{-1em}
\end{table}
\paragraph{Error Rate Analysis}
We look into the return type correctness for recursive code in three different settings: \textbf{Implicit-Dynamic Type(\texttt{gpt-3.5-turbo})}, where the return type can vary but is not specified in the question. \textbf{Explicit-Dynamic Type(\texttt{gpt-3.5-turbo})}, where the return type is explicitly specified in the question, and \textbf{Explicit-Dynamic Type(CodeLlama)} where we implement on the CodeLlama model. 

\begin{figure}
    \centering
\includegraphics[width=8cm]{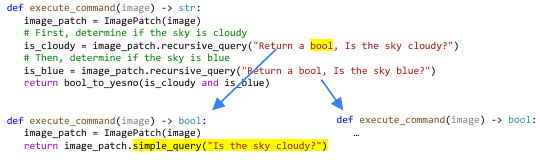}
    \caption{\textbf{CodeLlama Type Confusion Example.} This figure illustrates an instance where CodeLlama confuses types, despite having the return type explicitly specified. Even though the recursive\_query explicitly states ``Return a bool'' and generates a function that begins with the signature `-> bool', the function returns an incorrect type str.}
    \label{fig:codellama_error}
\end{figure}

\begin{figure}
    \centering
\includegraphics[width=8cm]{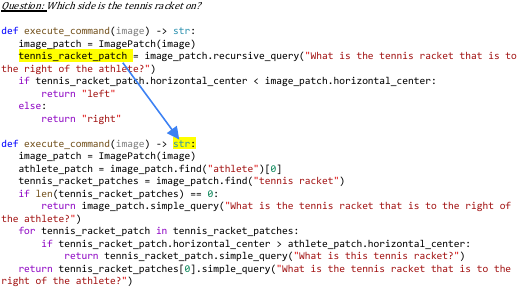}
    \caption{\textbf{Implicit Return Type Confusion Example.} This figure illustrates that without specifying the return type, the model often does not no what type to assign and begins the code with the wrong signature.}
    \label{fig:implicit_error}
\vspace{-10pt}
\end{figure}

We test on the GQA val\_2000 set. The return type error rates are presented in Table~\ref{tab:error_rate}. We find that with the return type specified, \texttt{gpt-3.5-turbo} model always returns the correct type. CodeLlama gets most of the return types correct but has 4.91\% errors. The errors are primarily due to type confusion. For instance, in Figure~\ref{fig:codellama_error}, despite the code specifying `Return a bool' and starting the generated code with the bool signature, the model still generates the incorrect type, str. This indicates that CodeLlama is not sensitive to tracking types within code to ensure the desired return type, which could be an area for future improvement.
In the case without explicit type specification, the model often fails to determine the appropriate type to generate and starts the code with a wrong signature, leading to unintended result. For example, in Figure~\ref{fig:implicit_error}, the code expects a return type ImagePatch but receives a str instead. This is very common error and results in a high error rate of 56.14\%.
\begin{table}[htbp]
\small
\centering
\caption{\textbf{Return Type Error Rate} We investigate the return type error rate under three settings. \texttt{gpt-3.5-turbo} consistently gets all the return types correct, aligning with the signature. In contrast, CodeLlama sometimes returns types that are not consistent with the given signature. Implicit types often lead to the code starting with an incorrect signature.}
\begin{tabular}{@{}lcc@{}}
\toprule
 & Return Type Error Rate & Acc\\
\midrule
CodeLlama(RVP-Explicit) & 4.91 & 49.27\\
gpt-3.5(RVP-Explicit) & \textbf{0}  &\textbf{54.75}\\
gpt-3.5(RVP-Implicit) & 56.14 & 52.74\\
\bottomrule
\end{tabular}
\label{tab:error_rate}
\vspace{-20pt}
\end{table}
\paragraph{Question Complexity Analysis}
We classify and test GQA problems by the number of logic steps (provided by the GQA metadata field \textit{relates}, e.g., What is the color of A that is to the right of B below C contains 2 steps of \textit{relates}) and the number of properties. 
\begin{table}[htbp]
\small
\centering
\caption{\textbf{Question Complexity Analysis.} RVP attains larger performamce gain over non-recursive method on more complicated questions.}
\begin{tabular}{@{}lcc@{}}
\toprule
 & Non-Recursive & Recursive \\
\midrule
\multicolumn{3}{c}{GQA-Relates}\\
\midrule
0-step & 61.34 & 62.50 (+1.16) \\
1-step & 37.11 & 44.40 (+7.29) \\
2-step & 32.37 & 40.29 (+7.92) \\
\midrule
\multicolumn{3}{c}{GQA-Properties}\\
\midrule
1-step & 64.62 & 62.45 (-2.17)\\
2-step & 54.84 & 59.67 (+4.83) \\
\bottomrule
\end{tabular}
\label{tab:complexquestions}
\vspace{-1em}
\end{table}
Thre results are shown in Table \ref{tab:complexquestions}. Results indicate that (i) As the problems get more complicated, both methods perform worse, and (ii) The recursive method mostly benefits questions that require more logical steps and properties. 

\paragraph{Other Prompting Methods} While our method is not a prompting method, we experimented with how the prompting methods in the NLP domain could help visual reasoning. Specifically, on the GQA\_val2000, we conducted experiments using Chain of Thought prompting\cite{wei2022chain}, Program of Though Prompting\cite{chen2022Program}, and Multi-subfunction prompting. 

For Chain of Thought prompting, we tried two methods: (i) adding a reasoning sentence, ``Let's think step by step, first we need to..."-like sentence before generating each code piece. We refer to this as CoT-before-code (ii) adding a ``Let's think step by step, first we need to..."-like sentence as the first comment inside each code piece. We refer to this as CoT-after-code. For Program of Thought, we followed the step-by-step code generation prompting.
For sub-functions prompting, we put all the sub-functions into a
single long function as the prompt and let the model generate a single code piece and define sub-functions inside it.

The results are shown in Table\ref{tab:prompting}. For CoT-before-code, we found that the model often fails to simultaneously generate the reasoning sentence and then the code, yielding in bad performance. For CoT-after-code, this problem doesn't exist but adding the long reasoning sentence doesn't help with the generated code. For PoT, the result is slightly better than CoT. And for sub-function generation, the result is better than CoT and PoT, but worse than RVP. This suggests that having modular functions could better help with visual programming than just having language reasoning, but long sequential code pieces hinder the model’s
capacity to capture the structure, as demonstrated in recent
studies\cite{liu2023lost}.

\begin{table}[htbp]
\small
\centering
\caption{\textbf{Comparison between different prompting methods}: Adopting the NLP prompting methods directly to aid visual programming does not yield in performance gain. It is not clear how these prompting methods can help visual reasoning out-of-box.}
\begin{tabular}{@{}lc@{}}
\toprule
 &  Accuracy \\
\midrule
CoT-before-code & 48.42\\
CoT-inside-code &  49.07\\
PoT & 51.53\\
Sub-functions & 52.03\\
RVP & \textbf{54.75}\\
\bottomrule
\end{tabular}
\label{tab:prompting}
\end{table}
\paragraph{In-Context Example Choice Study} In the main paper, we test on retrieval-based in-context example selection and fixed in-context example selection, we show the distribution of dynamic types using both the retrieval-based and non-retrieval-based methods in Table \ref{tab:retrieval_datatype_distribution}.
\begin{table}[htbp]
\small
\centering
\caption{\textbf{Diverse return types of embedding-based retrieval and fixed examples}}
\begin{tabular}{@{}lcc@{}}
\toprule
Data Type & Retrieval & Non-retrieval\\
\midrule
Str & 36 & 0\\
Bool & 188 & 123 \\
List[str] & 0 & 6 \\
ImagePatch & 178 & 1 \\
List[ImagePatch] & 2 & 1  \\
Float & 16 & 2 \\
\bottomrule
\end{tabular}
\label{tab:retrieval_datatype_distribution}
\vspace{-1em}
\end{table}
\section{Additional Visualization Results}
\label{sec:additional_examples}
\begin{figure*}
    \centering
\includegraphics[width=12cm]{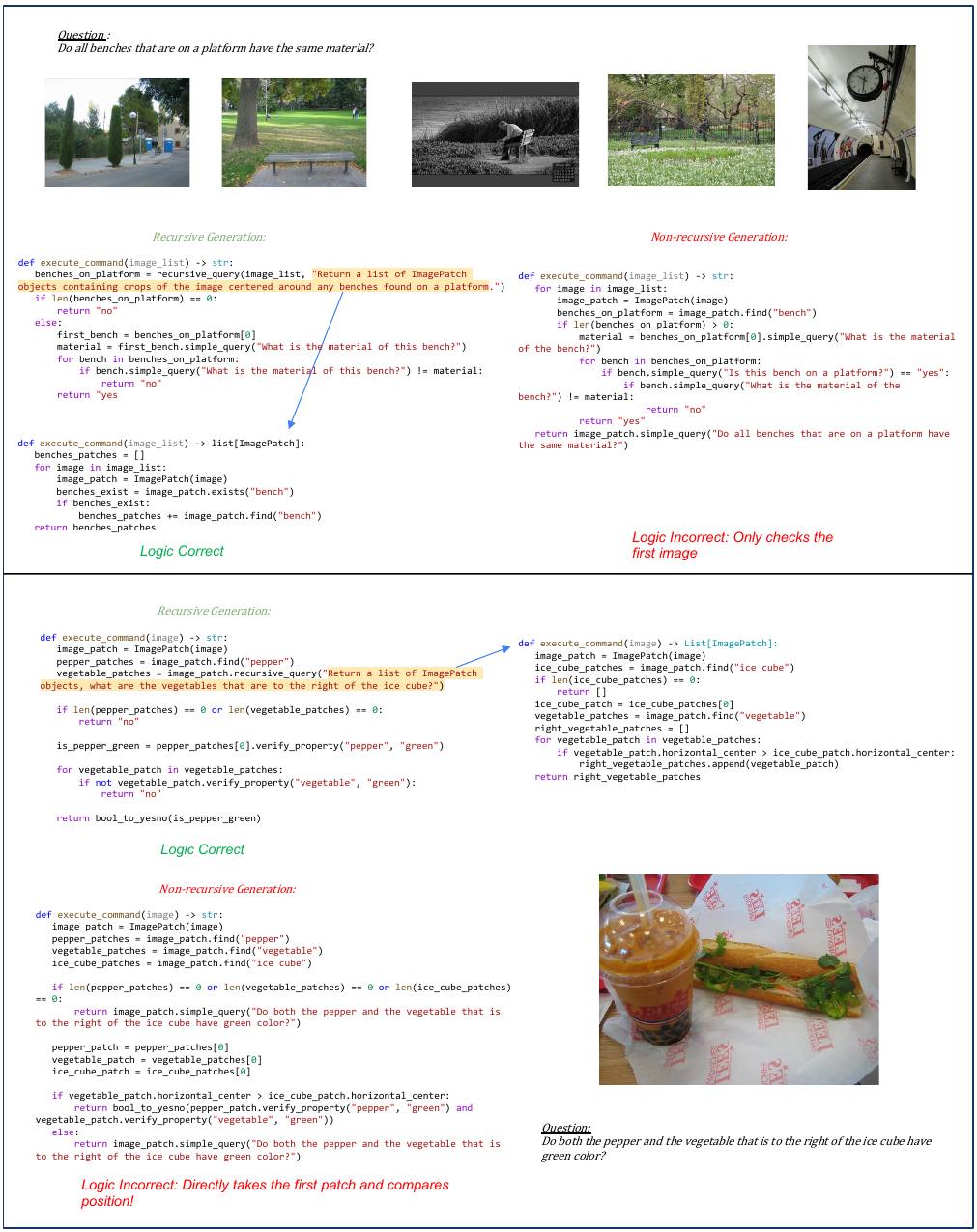}
    \caption{\textbf{Examples of COVR and GQA.}}
    \label{fig:example2}
\end{figure*}
\begin{figure*}
    \centering
\includegraphics[width=12cm]{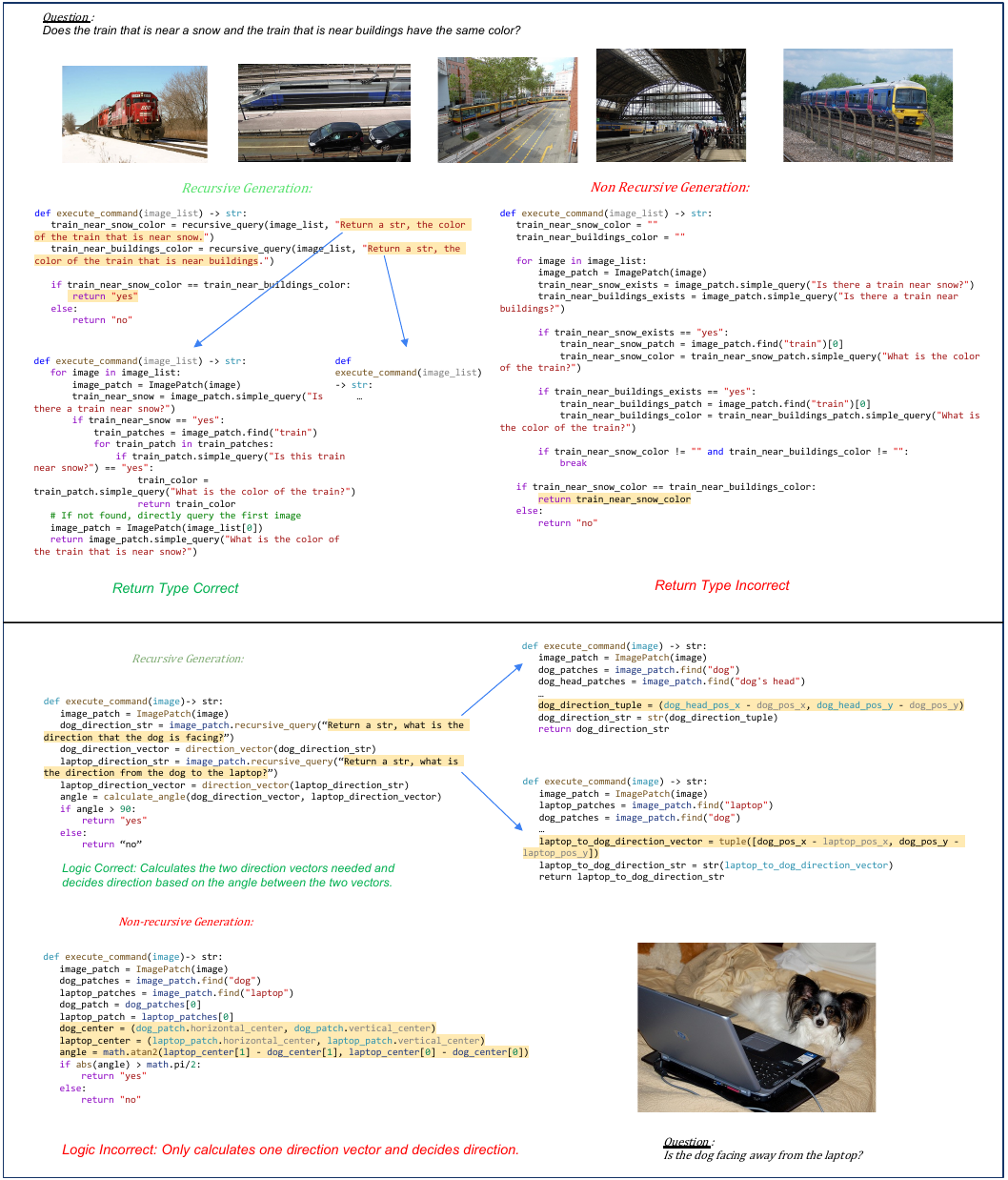}
    \caption{\textbf{Examples of COVR and VSR.}}
    \label{fig:example3}
\end{figure*}
\begin{figure*}
    \centering
\includegraphics[width=12cm]{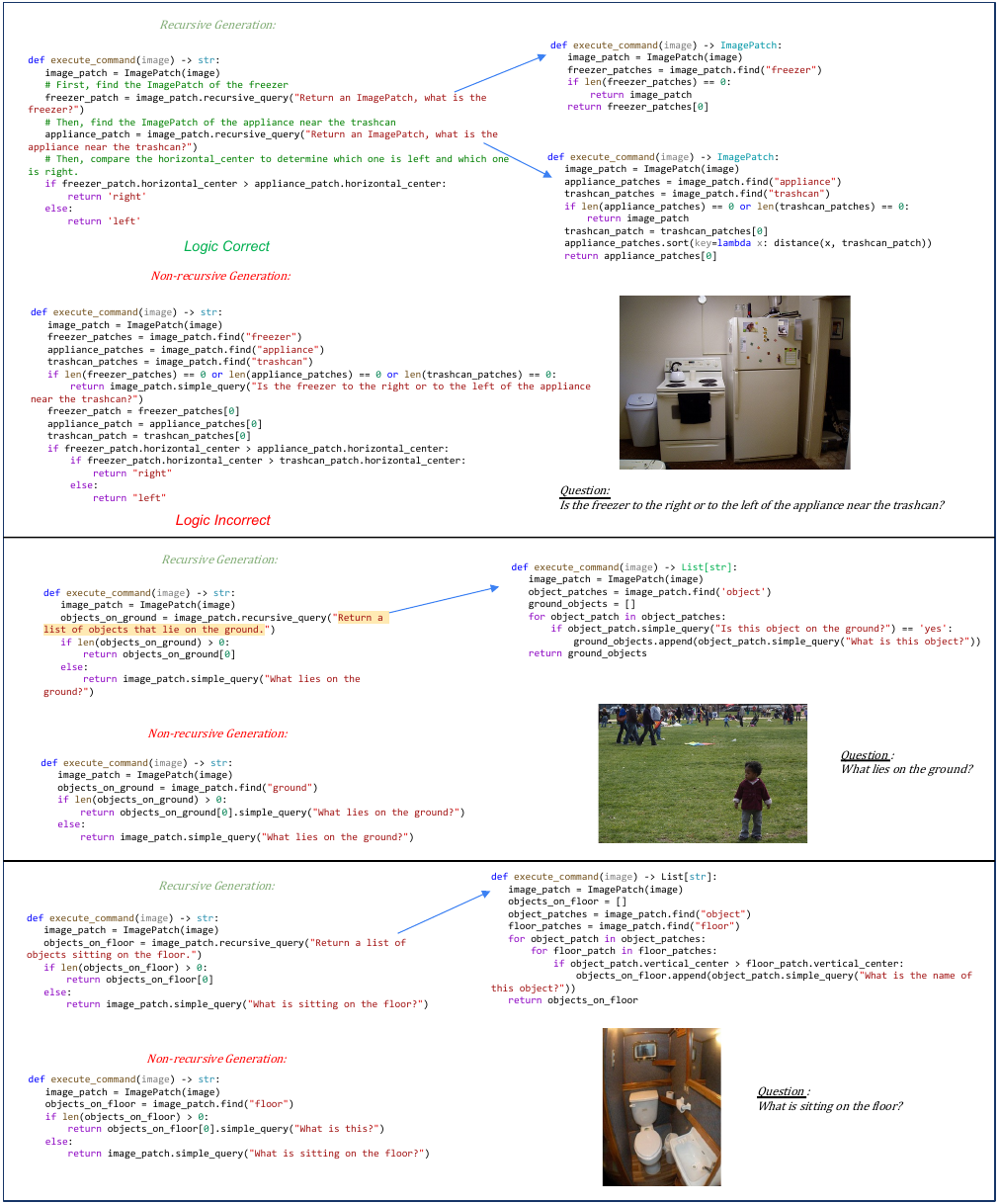}
    \caption{\textbf{Examples of GQA.}}
    \label{fig:example5}
\end{figure*}
In this section, we provide more visualization results for RVP. As shown in Figure~\ref{fig:example2},\ref{fig:example3},\ref{fig:example5}. In these visualization results, we show results from different datasets, diverse dynamic return types by RVP, and cases where RVP demonstrates correct logic flow, handles details more elegantly, or generates more readable codes.
\section{Additional Details}
In this section, we provide more details about the datasets we use. The model details between the model we use and previous models. Then some additional details of the survey we conduct.
\subsection{Dataset Details}
\paragraph{GQA}
GQA \cite{gqa} is a benchmark comprising compositional reasoning questions over real images. It includes a total of 22M questions, each of which can be represented by a functional program specifying the reasoning steps required to answer the question.
\paragraph{VSR} 
VSR \cite{vsr} is a spatial reasoning benchmark containing over 10k text-image pairs, featuring 66 types of spatial relations. As current Vision Language models struggle with spatial relationships, this benchmark poses a challenge in determining whether a spatial relation in an image is true or false.
\paragraph{COVR} 
COVR \cite{covr} focuses on visually-grounded compositional generalization with real images. It includes 262k examples based on 89k images, with 13.9k questions manually validated and paraphrased. The model must consider all images in a given list and conduct compositional reasoning to answer the question.
\paragraph{Next-QA} 
Next-QA \cite{nextqa} is a VideoQA benchmark for video question answering. We use the multi-choice version, where the model must choose the correct answer from 5 options. The ATP-T \cite{atp} split, comprising the more challenging questions in Next-QA that require complex temporal reasoning, includes around 900 examples in total.
\begin{figure*}[h]
    \centering
\includegraphics[width=12cm]{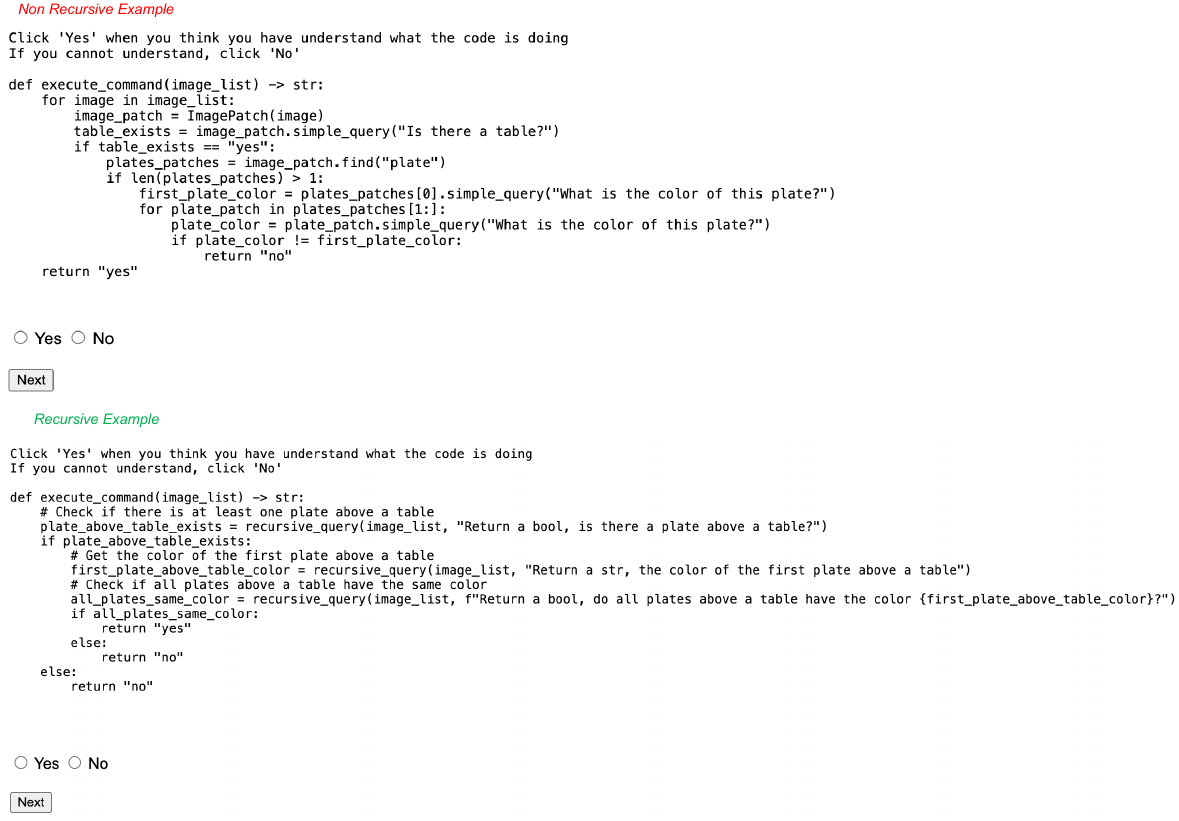}
    \caption{\textbf{Recursive and Non-recursive examples from the survey.} The participant is asked to read the code, try to understand what the code is doing as quickluy as possible, and then proceed by clicking on the button `Next'.}
    \label{fig:survey_screenshot}
\end{figure*}
\subsection{Model Details}
\paragraph{CodeLlama} 
CodeLlama models are based on Llama2 models \cite{touvron2023llama} and trained for code generation. They are open-sourced and have shown great coding abilities in various settings. 
The class includes foundation models (Code Llama), Python specializations (Code Llama - Python), and instruction-following models (Code Llama - Instruct), with versions containing 7B, 13B, and 34B parameters, respectively. In our experiments, we used CodeLlama-13b-Python-hf, which is relatively small for GPU compatibility and is specifically trained for generating high-quality Python code."
\paragraph{\texttt{code-davinci-002}} 
\texttt{code-davinci-002} is a coding model developed by OpenAI and the most powerful coding model before GPT-4. Previous visual programming methods such as \cite{vipergpt, codevqa} are based on this model. Built on GPT-3, it was trained on billions of lines of source code in addition to natural language. However, the model is currently unavailable.
\paragraph{\texttt{gpt-3.5-turbo}} 
In all of our experiments, we use \texttt{gpt-3.5-turbo} as the model. This model improves over the original GPT-3 architecture, making it particularly effective for real-time applications. \texttt{gpt-3.5-turbo} is widely used for tasks that require quick and accurate natural language understanding and generation. Despite its capabilities, \texttt{gpt-3.5-turbo} is not specifically trained on code, which can lead to buggy code generation. It also shows limitations in robust in-context learning \cite{ye2023comprehensive}.
\subsection{Survey Details}
Our criterion for selecting survey participants is a background in computer science and the ability to easily understand Python code. Specifically, we distribute the survey link to undergraduate and graduate students in computer science programs. The first question in the survey is a simple, unused question designed to help participants become familiar with the survey format. An example screenshot of a survey question we show to the participants is displayed in Figure~\ref{fig:survey_screenshot}.
\begin{figure*}[h]
    \centering
    \includegraphics[width=\textwidth]{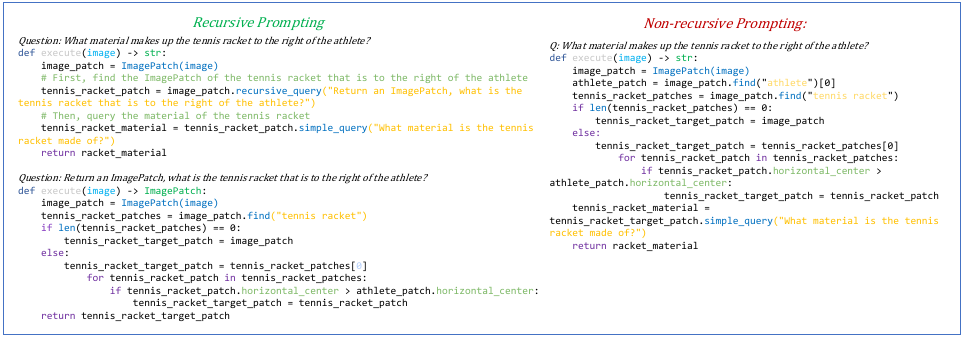}
    \caption{\textbf{The in-context examples we used for RVP and non-recursive method.} In the non-recursive case, the example is structured as a question followed by a single piece of code that contains all the logic and details. In RVP, we decompose the same non-recursive code into several pieces, linked by recursive\_query calls.}
    \label{fig:prompt_pipeline}
\end{figure*}
\subsection{In-Context Examples} 
We show an example of the in-context examples we used between RVP and ViperGPT in Figure~\ref{fig:prompt_pipeline}. In the non-recursive case, the example is structured as a question followed by a single piece of code that contains all the logic and details. In RVP, we decompose the same non-recursive code into several pieces, linked by recursive\_query calls.
\end{document}